\documentclass[sigconf]{acmart}

\usepackage{booktabs} %
\usepackage{subcaption}
\usepackage{multirow}
\usepackage{tabularx}
\usepackage[ruled]{algorithm2e} %

\SetAlFnt{\small}
\SetAlCapFnt{\small}
\SetAlCapNameFnt{\small}
\SetAlCapHSkip{0pt}

\acmJournal{TOG}

\copyrightyear{2022}
\acmYear{2022}
\setcopyright{acmlicensed}\acmConference[SIGGRAPH '22 Conference Proceedings]{Special Interest Group on Computer Graphics and Interactive Techniques Conference Proceedings}{August
7--11, 2022}{Vancouver, BC, Canada}
\acmBooktitle{Special Interest Group on Computer Graphics and Interactive Techniques Conference Proceedings (SIGGRAPH '22 Conference Proceedings), August 7--11, 2022, Vancouver,
BC, Canada}
\acmPrice{15.00}
\acmDOI{10.1145/3528233.3530748}
\acmISBN{978-1-4503-9337-9/22/08}

\newcommand{\Fig}[1] {Fig.~\ref{fig:#1}}

\newcommand{\Sec}[1] {Sec.~\ref{sec:#1}}

\newcommand{\Etal}   {{\textit{et~al.}}}

\begin{document}
\title{Deep Deformable 3D Caricatures with Learned Shape Control}

\author{Yucheol Jung}
\orcid{0000-0003-1593-4626}
\affiliation{%
  \institution{POSTECH}  
  \city{Pohang}
  \country{Republic of Korea}
  }
\email{ycjung@postech.ac.kr}

\author{Wonjong Jang}
\orcid{0000-0002-1442-9399}
\affiliation{%
  \institution{POSTECH} 
  \city{Pohang}
  \country{Republic of Korea}
  }
\email{wonjong@postech.ac.kr}

\author{Soongjin Kim}
\orcid{0000-0001-8142-7062}
\affiliation{%
  \institution{POSTECH} 
  \city{Pohang}
  \country{Republic of Korea}
  }
\email{kimsj0302@postech.ac.kr}

\author{Jiaolong Yang}
\orcid{0000-0002-7314-6567}
\affiliation{%
  \institution{Microsoft Research Asia}
  \city{Beijing}
  \country{China}
}
\email{jiaoyan@microsoft.com}

\author{Xin Tong}
\orcid{0000-0001-8788-2453}
\affiliation{%
  \institution{Microsoft Research Asia}
  \city{Beijing}
  \country{China}
}
\email{xtong@microsoft.com}

\author{Seungyong Lee}
\orcid{0000-0002-8159-4271}
\affiliation{%
  \institution{POSTECH}
  \city{Pohang}
  \country{Republic of Korea}
}
\email{leesy@postech.ac.kr}

\begin{abstract}

A 3D caricature is an exaggerated 3D depiction of a human face.
The goal of this paper is to model the variations of 3D caricatures in a compact parameter space so that we can provide a useful data-driven toolkit for handling 3D caricature deformations.
To achieve the goal, we propose an MLP-based framework for building a deformable surface model, which takes a latent code and produces a 3D surface. 
In the framework, a SIREN MLP models a function that takes a 3D position on a fixed template surface and returns a 3D displacement vector for the input position.
We create variations of 3D surfaces by learning a hypernetwork that takes a latent code and produces the parameters of the MLP.
Once learned, our deformable model provides a nice editing space for 3D caricatures, supporting label-based semantic editing and point-handle-based deformation, both of which produce highly exaggerated and natural 3D caricature shapes. 
We also demonstrate other applications of our deformable model, such as automatic 3D caricature creation. Our code and supplementary materials are available at \href{https://github.com/ycjungSubhuman/DeepDeformable3DCaricatures}{https://github.com/ycjungSubhuman/DeepDeformable3DCaricatures}.

\end{abstract}

\begin{teaserfigure}
\includegraphics[width=\textwidth]{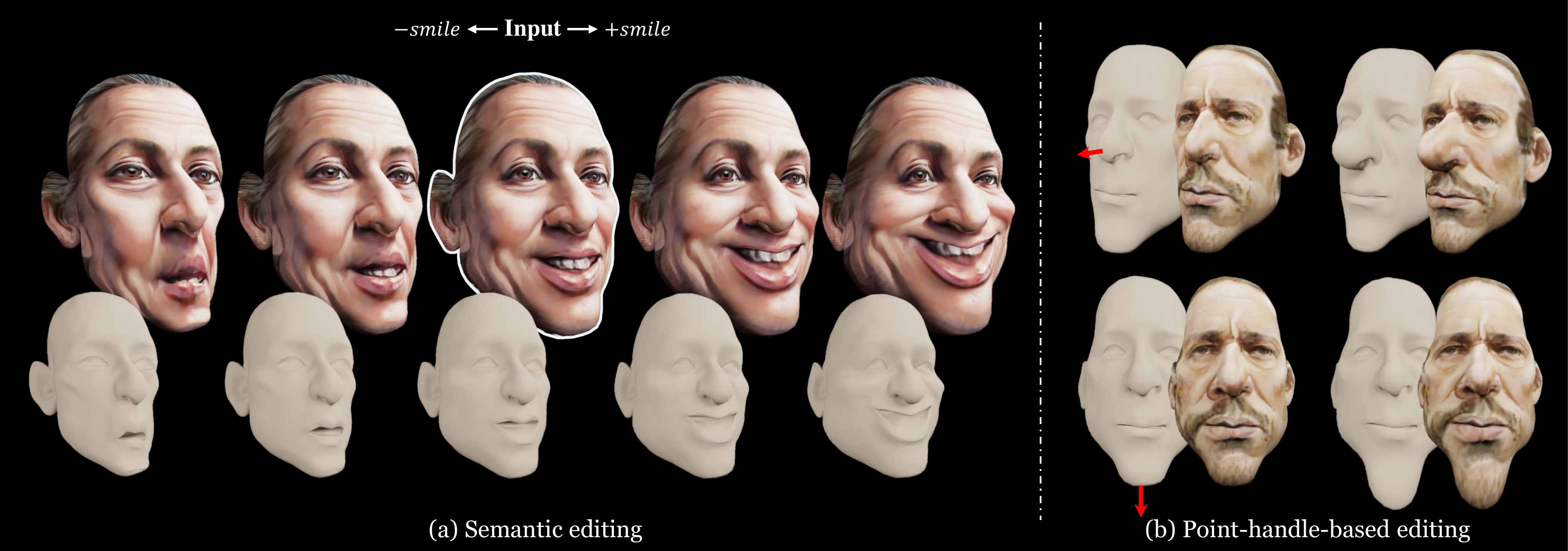}
\caption{Our deformable model provides a data-driven editing space for 3D caricature shapes. (a) Semantic editing of 3D caricatures with varying degrees. (b) Point-handle-based deformation using learned latent space. The edited 3D caricatures have been reconstructed from 2D caricatures generated by StyleCariGAN \cite{jang2021stylecarigan}. 3D caricatures can also be reconstructed from real-world 2D caricatures, e.g., created by artists.
}
\label{fig:teaser}
\end{teaserfigure}

\ccsdesc[500]{Computing methodologies~Mesh models}
\ccsdesc[500]{Computing methodologies~Neural networks}

\keywords{Deformable model, Parametric model, 3D face model, Semantic 3D face control, 3D face deformation, Auto-decoder}

\maketitle

\section{Introduction}

Deformable models have been widely used in graphics for shape creation and control.
They provide compact parameter spaces and \textit{a priori} information for the parameters, enabling effective manipulation of shapes.
Human face is one of the domains that benefits from deformable models. One of the most widely used formulation for human face is the 3D Morpahble Model (3DMM) \cite{blanz1999morphable}. 3DMM and its variants have been adopted for challenging tasks, including 3D face reconstruction from 2D images \cite{deng2019accurate, garrido2016reconstruction}, semantic 3D face editing \cite{partbased}, and face reenactment \cite{thies2016face2face}.

A 3D caricature is an exaggerated 3D depiction of a human face. 
It has broad applications ranging from professional cartoon film production to low-cost avatar creation for games, social media, and AR/VR.
Conventionally, 2D caricatures are created by skilled artists, and crafting 3D caricatures would require even more expertise.
On the other hand, similarly to the case of 3D faces, a deformable 3D caricature model can provide a useful toolkit for handling exaggerated 3D faces by modeling the variations of 3D caricatures (\Fig{teaser}).

3DMM has been used for effectively modeling variations of regular faces with a 
linear parameter space for 3D vertex positions. When applied to 3D caricatures, 3DMM results in large reconstruction errors as shown in a previous work \cite{wu2018alive}. 
To reduce the reconstruction errors, Wu \Etal~\shortcite{wu2018alive} proposed an explicit non-linear mapping from shape parameters to 3D vertex positions for extrapolating regular faces to 3D caricatures. 
However, they did not exploit at all the editing capability of the deformable model defined by the mapping.
Recently released 3DCaricShop dataset \cite{qiu20213dcaricshop} provides a set of high-quality 3D caricature meshes sculpted by 3D artists.
Such a dataset enables learning a latent space together with the mapping to 3D caricatures automatically using a neural network. %
Still, learning from the dataset is challenging as 3D caricatures have highly diverse styles, yet the number of examples is limited to around 2K. We provide an analysis on the complexity of the dataset in the supplementary material.

Our goal is to learn a deformable model from 3DCaricShop dataset \cite{qiu20213dcaricshop} to create a toolkit for controllable 3D caricatures. 
With highly diverse and sparse data, we believe the network architecture plays a central role in the learning.
We choose a multi-layer perceptron (MLP) for the network architecture, which has been successfully used for various applications, including shape reconstruction \cite{park2019deepsdf, saito2019pifu} and neural rendering \cite{mildenhall2020nerf, lombardi2018deep, lombardi2019neural, chan2021pi}.

\begin{figure}[t]
    \centering
    \small
    \begin{tabularx}{\textwidth}{ccc}
    \includegraphics[width=0.13\textwidth]{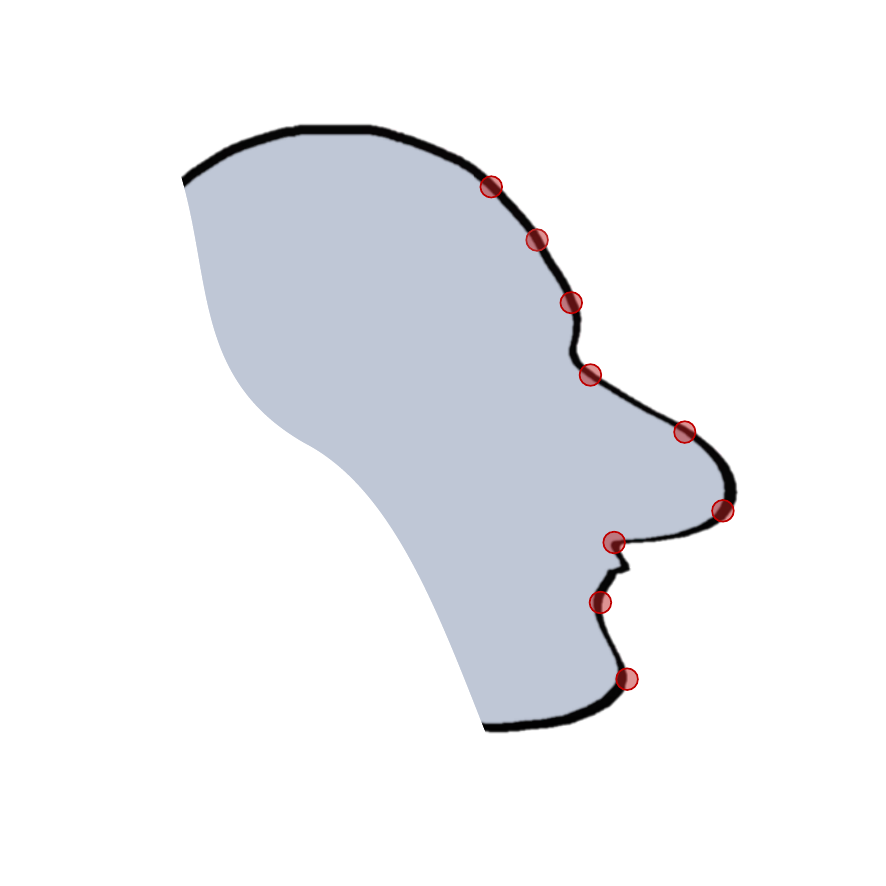} & 
    \includegraphics[width=0.13\textwidth]{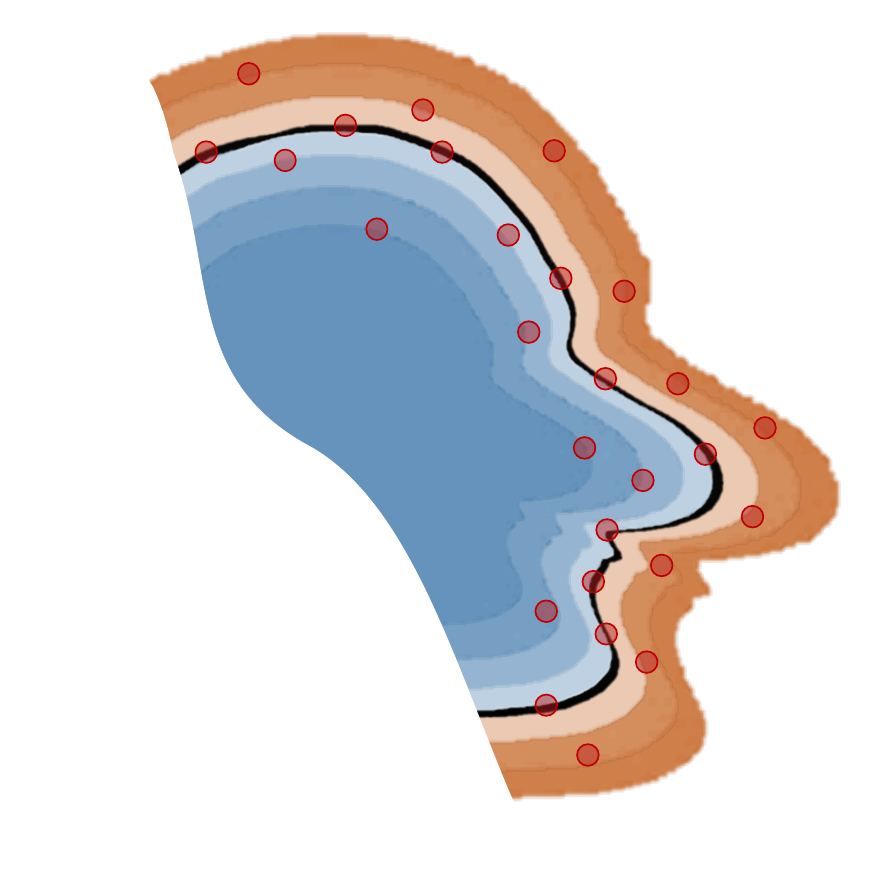} & 
    \includegraphics[width=0.13\textwidth]{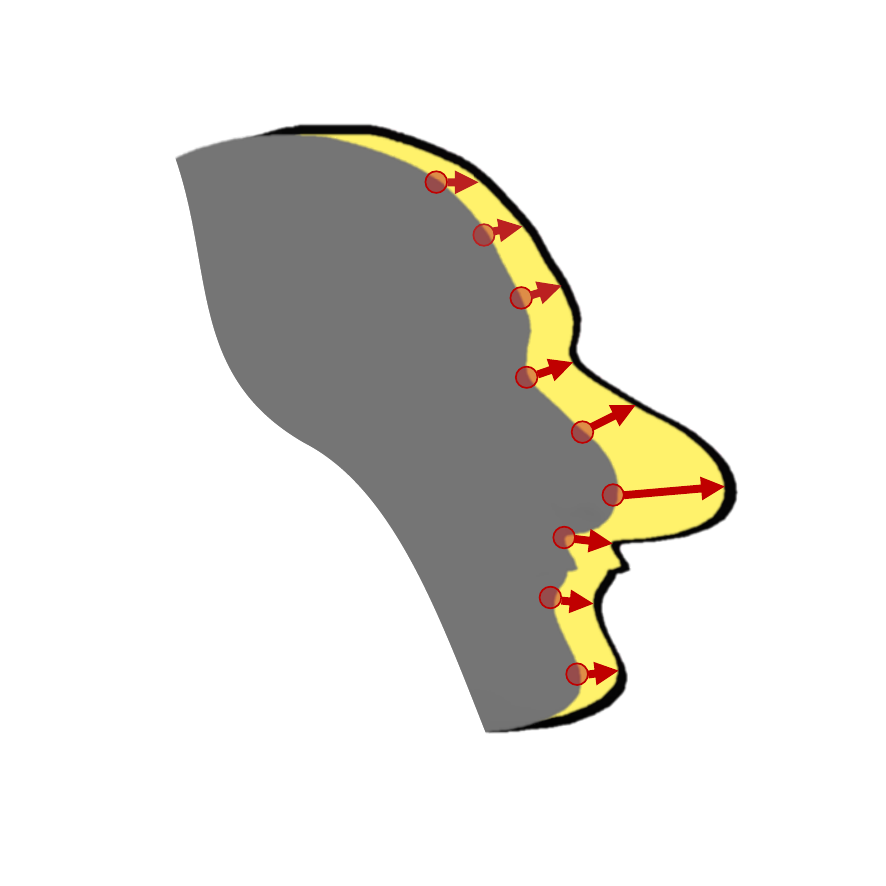} \\
    (a) Vertex position & (b) SDF & (c) Deformation
    \end{tabularx}
    \centering
     \caption{Different ways to represent a 3D shape on a network. }
    \label{fig:volume-vs-surface}
\end{figure}

Another important decision for the learning is the output representation of the MLP
(\Fig{volume-vs-surface}). An array of vertex position is a straightforward approach when all dense correspondence is given in the dataset. However, we observed 
an MLP for generating vertex array suffers from slow convergence and high reconstruction error. This approach also limits the sampling positions on the surface only to vertices. Continuous SDF in a volume is widely used recently for general 3D shape modeling, but we observed an MLP for SDF tends to miss small details such as the shape of eyes. Consequently, our framework lets the MLP to model continuous deformation function on a template surface. With this approach, we can generate 3D caricatures with important details.

To this end, we propose an MLP-based framework for building a deformable surface model, which takes a 128-dimension latent code and produces a 3D surface. In our framework, a SIREN MLP \cite{sitzmann2020implicit} models a function that takes a 3D position on a fixed template surface and returns a 3D displacement vector for the input point. 
We create variations of 3D surface by learning a hyper-network that takes a latent code and produces the parameters of the MLP. The MLP and the hyper-network are learned jointly to build an auto-decoder \cite{park2019deepsdf}, which is used as our deformable 3D surface model. 

Once learned, our deformable model provides a nice editing space for 3D caricatures. We demonstrate label-based semantic 3D caricature editing and point-handle-based 3D caricature editing, both of which produce highly exaggerated and natural 3D caricature shapes.
We also demonstrate other applications, such as automatic 3D caricature creation from a 3D or 2D regular face and fitting to 2D landmarks to reconstruct 3D geometry out of a 2D caricature.

The contributions of this paper can be summarized as follows;
\begin{itemize}
    \item We propose an MLP-based framework for building a deformable model that captures the variations of 3D caricatures in a compact parameter space.
    \item Our deformable model provides a useful data-driven toolkit for handling 3D caricatures, supporting label-based semantic editing and point-handle-based deformation as well as other applications.
\end{itemize}

\section{Related Work}

\begin{figure*}
    \centering
    \includegraphics[width=0.9\textwidth]{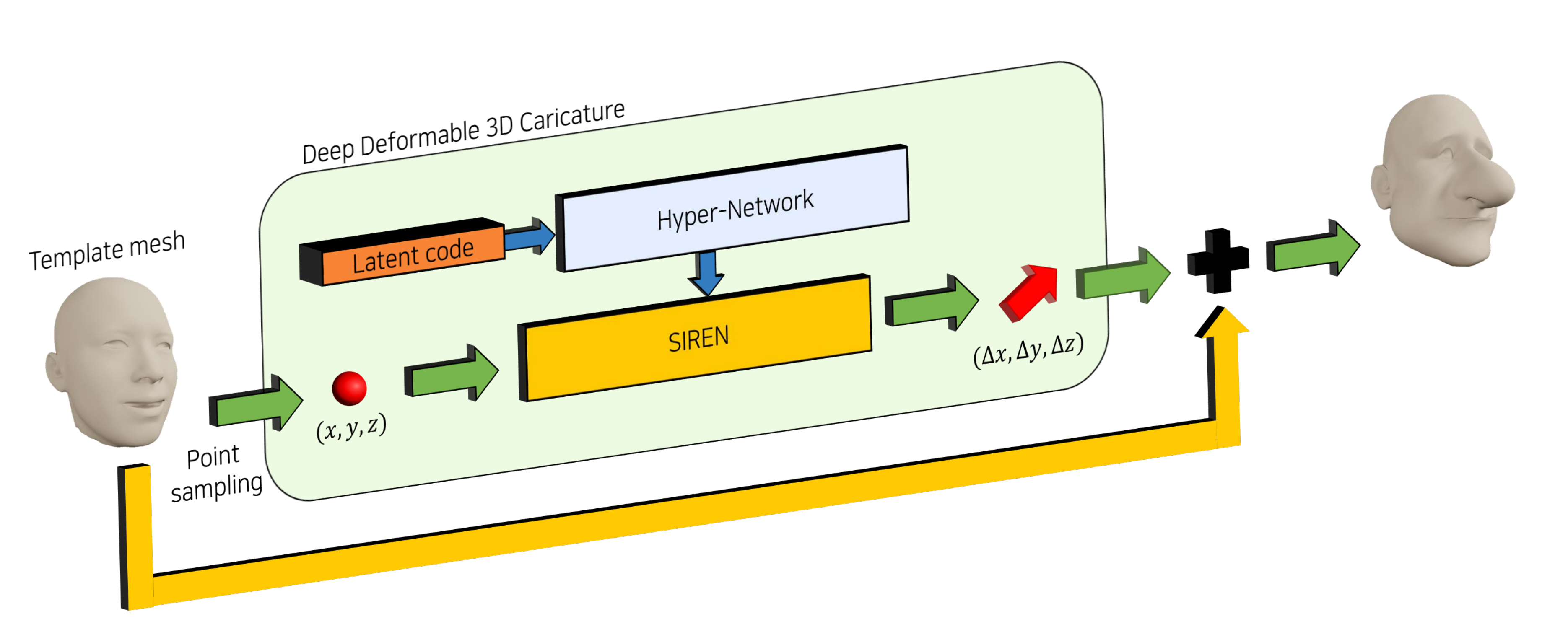}
    \caption{Overall framework for our deep deformable 3D caricature model.}
    \label{fig:overview}
\end{figure*}
\subsection{Deformable face model}
3DMM \cite{blanz1999morphable} introduces a linear subspace for shape and texture using PCA decomposition from a 3D face database. Wu~\Etal~\shortcite{wu2018alive} showed 3DMM is not suitable for defining a subspace for caricatures since the space is not large enough to model large variations in caricatures. Different formulations on the 3DMM have been studied \cite{luthi2017gaussian, ploumpis2020towards}, but their implication on the 3D caricature domain has not been explored.

Deep learning-based methods for modeling 3D facial mesh has been studied recently. PRNet \cite{feng2018joint} uses a 2D image to represent 3D facial geometry. %
Lombardi~\Etal~\shortcite{lombardi2018deep} builds a variational auto-encoder of 3D face meshes using fully connected layers. CoMA \cite{ranjan2018generating} proposes an auto-encoder model using graph convolution together with specialized down-sampling and up-sampling operators.
MeshGAN \cite{cheng2019meshgan} trains two CoMA-based mesh decoders for facial identity and expression. Jiang~\Etal~\shortcite{jiang2019disentangled} disentangles 3D facial meshes into identity and expression, both of which are encoded by graph-convolution-based mesh auto-encoders. In contrast to these methods generating fixed set of discrete vertices, we propose to model shape variations as a continuous deformation function using MLP to build a deformable 3D caricature model.

\subsection{Deformable caricature model}
One approach for building a deformable 3D caricature model is to expand regular face dataset with computational exaggeration. DeepSketch2Face \cite{han2017deepsketch2face} extends regular face dataset by exaggerating faces synthetically using \cite{sela2015computational} to build a bi-linear model of 3D caricatures. CaricatureShop \cite{han2018caricatureshop} uses synthetic 3D caricatures as training data. DeepSketch2Face and CarictureShop only focuses on sketch-based 3D caricature editing. Alive Caricature \cite{wu2018alive} models a space of 3D caricatures by building a carefully-designed blending operations for regular 3D faces. 3DCaricShop \cite{qiu20213dcaricshop} proposes a 3D caricature dataset sculpted by artists, and use the 3D caricatures to build a PCA model, which is used as an artifact-filtering tool. 
Alive Caricature and 3DCaricShop focuses on 3D mesh reconstruction from 2D caricatures through free-form vertex deformation using an optimization guided by parametric models. Although these solution produces 3D caricatures faithful to input images, editing on the result is non-trivial since the result is not fully represented in an editable parameter space. In contrast, our data-driven parametric model for 3D caricatures provides readily available controls on the result when applied to 3D caricature reconstruction.

Cai~\Etal~\shortcite{cai2021landmark} learn caricature landmark detection and 3D caricature reconstruction using the results of Alive Caricature as training data. 3DMagicMirror \cite{guo20193d} trains a graph-convolutional variational auto-encoder using regular 3D faces and the results of Alive Caricature. The latent space is used for mapping a regular 3D face to a 3D caricature. These methods focus on specific tasks of landmark detection and automatic caricature creation. We provide a controllable parametric 3D caricature model that can be generally used for various sub-tasks including semantic editing.

\subsection{Deep-learning-based 3D shape model}

DeepSDF \cite{park2019deepsdf} proposed using MLP to learn SDF to represent a 3D shape using an auto-decoder framework. Following DeepSDF, effective learning of SDFs has been researched. Implicit geometric regularization \cite{gropp2020implicit} is widely used to regularize networks for SDF. DIF-Net \cite{deng2021deformed} learns a template SDF and volumetric deformation function using MLPs, providing dense correspondence between generated SDFs. %
AtlasNet \cite{groueix2018papier} represents a 3D shape as multiple surface patches in 2D domain to model the surfaces of general 3D objects.
In contrast, our method represents a 3D shape as a deformed template mesh.

3DN \cite{wang20193dn} reconstructs a 3D mesh from an input image by deforming a given fixed template mesh using MLPs. The function modeled by the MLP is similar to our framework; Both approaches model a continuous surface deformation function of a template surface. 3DN only discusses the effectiveness of the network in the context of 3D reconstruction via deformation of a given template. We use template deformation to effectively learn a controllable latent space of 3D shapes when dense correspondence between the shapes is given.

\section{Deep Deformable 3D Caricatures}

We build a data-driven deformable model for 3D caricatures using an auto-decoder framework \cite{park2019deepsdf}. Once learned, the latent code can be sampled from distribution or optimized for representing an arbitrary input, e.g., unseen 3D caricature or sparse landmarks. Various controls can be achieved using the learned latent space. \Fig{overview} shows our overall framework.

\subsection{3D caricature representation}

The key design choice in the training of the auto-decoder is the representation of a 3D caricature.  While other choices may exist, we discuss three possible representations: vertex position array, signed distance function (SDF), and surface deformation function, where we choose surface deformation function.

\paragraph{Vertex position array} Since the training data comes in the form of 3D mesh with uniform connectivity, one of the simplest approaches to model the data is to represent a shape as an array of per-vertex 3D position. Using this representation, we can generate a 3D caricature from a latent code by mapping the code to a large array of size $V \times 3$ using an MLP, where $V$ is the number of vertices. %
We observe this architecture suffers from slow convergence and high reconstruction error (See \Sec{comparison}).

\paragraph{SDF} 
DeepSDF \cite{park2019deepsdf} and its variants have been widely used for modeling general shapes with large and diverse variations, even including topology changes. However, MLPs for SDF result in inefficiency when applied to our case of 3D caricature, where dense vertex correspondence between meshes is provided in the dataset. In terms of applications, it is not trivial to retrieve uniform meshing and correspondence from the generated SDFs. UV parameterization may also not be trivially shared between generated 3D caricatures. Also, the MLPs for SDF should produce valid function values outside of the surface, while our region of interest is only on the surface shape. When we trained DeepSDF with the 3D caricature dataset, we observed a loss of crucial details such as the shape of the eyes (See \Sec{comparison}). 

\paragraph{Surface deformation}
We find modeling each shape as a deformation of a fixed template surface effective for learning the latent space of 3D caricatures, compared to using absolute position. We model deformation for each shape as a continuous function defined on the template surface. The continuous function is represented as an MLP and the variation in the function is modeled using the \textit{hyper-network} \cite{david2017hypernetworks} framework, which has recently been used in modeling variations in 3D shapes using continuous functions \cite{sitzmann2019scene, liu2020neural, deng2021deformed}. The hypernetwork architecture enables fast convergence, and the continuous representation enables finer sampling on the surface (See \Sec{continuous}). For a more detailed analysis of the convergence of hypernetwork, refer to \Sec{ablation-structure}.

Compared to using SDFs, our representation fully utilize the correspondence information in the dataset. Our representation has three advantages over an MLP for SDF: 1) Initial meshing and the UV parameterization is shared trivially among all generates shapes. 2) Unlike SDF defined on volume, the template deformation function is only required to have valid values on the template surface, so the network capacity can focus on a smaller domain. 3) The template contains essential components of human face, so the network can focus only on the variations of shapes and skip the learning to produce overall shapes of facial components.

\begin{figure}
    \centering
    \begin{tabular}{c@{}c@{}c@{}c}

    \includegraphics[width=0.11\textwidth, height=0.11\textwidth]{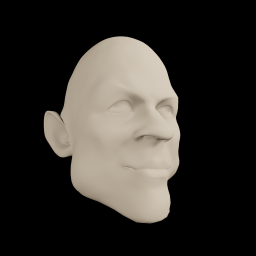} &
    \includegraphics[width=0.11\textwidth, height=0.11\textwidth]{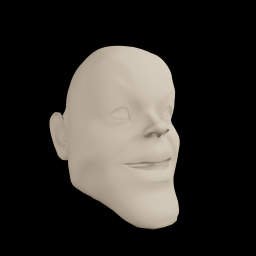} &
    \includegraphics[width=0.11\textwidth, height=0.11\textwidth]{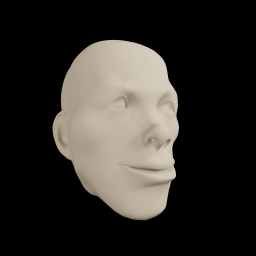} &
    \includegraphics[width=0.11\textwidth, height=0.11\textwidth]{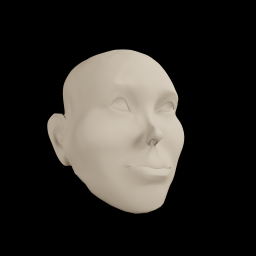} \\

    \end{tabular}
    \caption{Random 3D caricatures sampled from our deformable model.}
    \label{fig:randomsample}
\end{figure}

In conclusion, we represent a 3D caricature as a continuous displacement function defined on a fixed template surface. The function is modeled by an MLP that maps a 3D position sampled on the template surface to a 3D displacement vector to be applied to the input point. To generate a 3D caricature from a latent code, we map a latent code to the parameters of the MLP using a hypernetwork. The trained network can model diverse 3D caricatures with various expressions (\Fig{randomsample}).

\subsection{Network structure}

To learn the latent space of 3D caricature deformations, we adapt the \textit{Deform-net} network architecture of DIF-Net \cite{deng2021deformed}, which was used to deform and edit signed distance functions. Originally, the network takes a 3D position in volume to produce deformation for the coordinate to be fed into SDF. We reformulate the network by changing the goal 
to learning 
surface deformation defined on a fixed surface. Our network consists of two modules: SIREN MLP \cite{sitzmann2020implicit} for mapping a point on the template to a displacement vector and a hypernetwork that introduces variations in shapes by generating parameters for the SIREN MLP. We use the SIREN network since it converges fast and handles large shape variations effectively, as demonstrated in \cite{deng2021deformed}. 

A possible alternative to the hypernetwork is simply concatenating the latent code to the input and intermediate features as in \cite{park2019deepsdf, schwarz2020graf}. We observed hypernetwork provides faster convergence. The conditioning via latent code concatenation took much longer time to converge. More details on our network structure and comparison between the latent code concatenation and the hypernetwork are in \Sec{ablation-structure}.

\subsection{Network training}

\paragraph{Dataset}

We use 3DCaricShop dataset \cite{qiu20213dcaricshop}. The dataset contains 2K meshes sculpted by 3D artists. We obtained 1,409 registered 3D caricature meshes from the authors. The registered meshes have similar vertex connectivity as FaceWarehouse \cite{cao2013facewarehouse}, except the neck area is removed, and the holes in the eyes and the mouth are closed. We used 1,268 meshes as the training set, 14 meshes as the validation set, and 127 meshes as the test set. 

Along with the dataset, we obtained the mean face of FaceWarehouse that has the vertex connectivity of 3DCaricShop. We use the mean face as our template mesh. The mean face has an open mouth to avoid self-intersection around the lips. A template with an open mouth is important. If the template shape has closed mouth, since two different points on the upper lip and the lower lip have almost identical coordinates, mouth opening is impossible. With our template mesh, the mouth opens well with no problem.

\paragraph{Point sampling}

To train the network, we need point samples on the template mesh and its corresponding displacements. A straightforward approach is to use the vertices of the template mesh. The vertices contain important samples on locations where details are required, such as eyes. 
On the other hand, triangles on the template mesh have different sizes, and the shapes around large triangles, e.g., the shape of the cheek, can be more accurately captured by uniform random sampling on the surface. 
We use a mixture of these two approaches. Details on the point sampling algorithm are described in the supplementary material.

Our hybrid approach enables the trained network to produce more accurate shapes around large triangles compared to using vertices only. When the test set reconstruction error is calculated using uniform samples on the surface to take into account different triangle areas, the hybrid approach showed lower error of 0.0171, compared to 0.0188 of the vertices-only case. %

\paragraph{Loss}

We train the ReLU hypernetwork $H$ and the $M$-dimension latent code $z_k$ simultaneously, similarly to \cite{park2019deepsdf}. $z_k$ is initialized randomly using a zero-centered normal distribution of standard deviation 0.01 for each $k$-th training sample. A SIREN MLP $S$ takes a template 3D position and hypernetwork-generated parameters $H(z_k)$. The training is done using the loss
\begin{equation}
L(p, \hat{p}, z_k) = \dfrac{\lambda_{mse}}{N} \sum_i^N || p_i - S(\hat{p}_i, H(z_k)) - \hat{p}_i ||_2^2 + \dfrac{\lambda_{reg}}{M} || z_k ||_2^2,
\end{equation} where $p$ denotes the points on an example surface, $\hat{p}$ denotes the corresponding points on the template surface, $\lambda_{mse}$ controls the importance of data term, $\lambda_{reg}$ controls the regularization on the latent code, $N$ is the number of point samples. We use $\lambda_{mse}=3.0 \times 10^{3}$, $\lambda_{reg}=1.0 \times 10^{6}$, $N=23132$, and $M=128$ for training the network.

\section{Experiments}

\subsection{Comparison between different shape representations}

\paragraph{Vertex position array} 

\label{sec:comparison}

\begin{figure}
    \centering
    \small
    \renewcommand{\arraystretch}{0}\begin{tabular}{c@{}c@{}c}
    
    \raisebox{-.5\height}{\includegraphics[width=0.13\textwidth, height=0.13\textwidth]{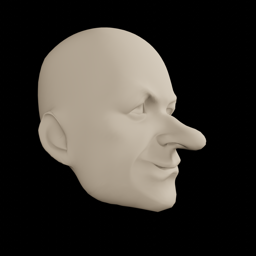}} &
    \raisebox{-.5\height}{\includegraphics[width=0.13\textwidth, height=0.13\textwidth]{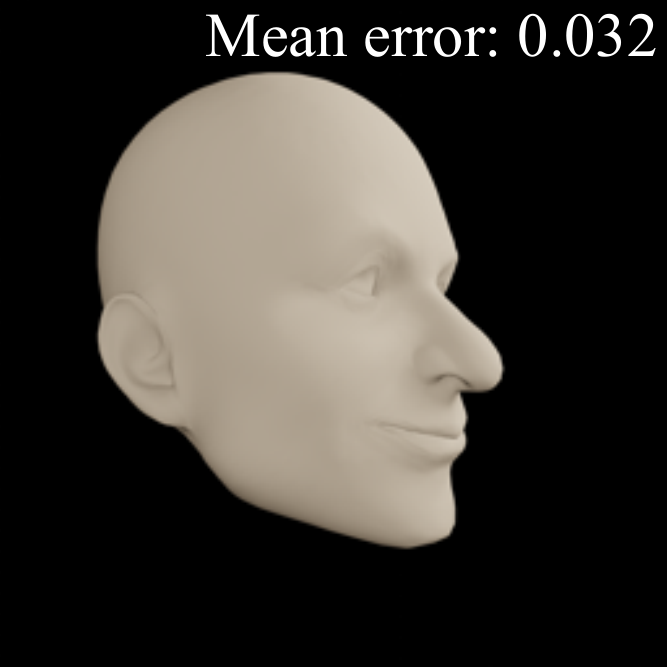}} &
    \raisebox{-.5\height}{\includegraphics[width=0.13\textwidth, height=0.13\textwidth]{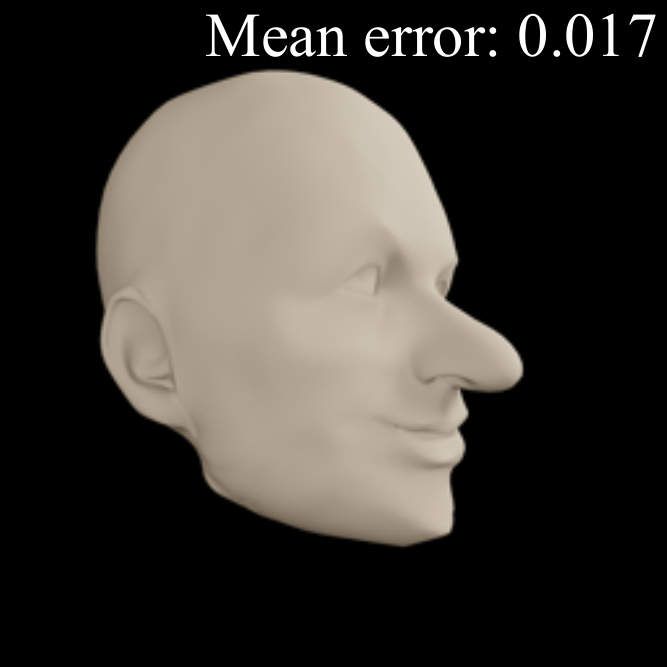}} \\
    
    \raisebox{-.5\height}{\includegraphics[width=0.13\textwidth, height=0.13\textwidth]{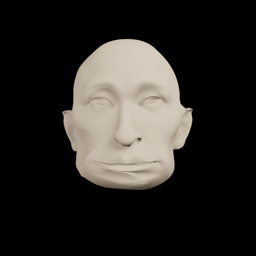}} &
    \raisebox{-.5\height}{\includegraphics[width=0.13\textwidth, height=0.13\textwidth]{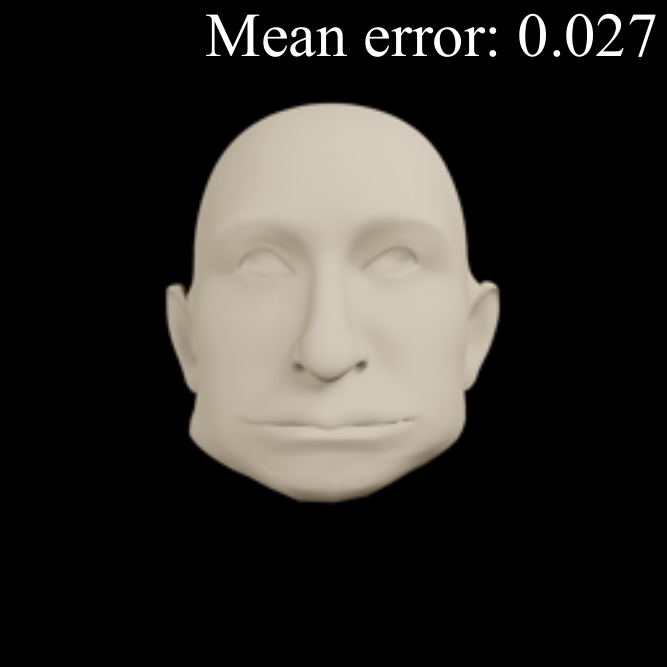}} &
    \raisebox{-.5\height}{\includegraphics[width=0.13\textwidth, height=0.13\textwidth]{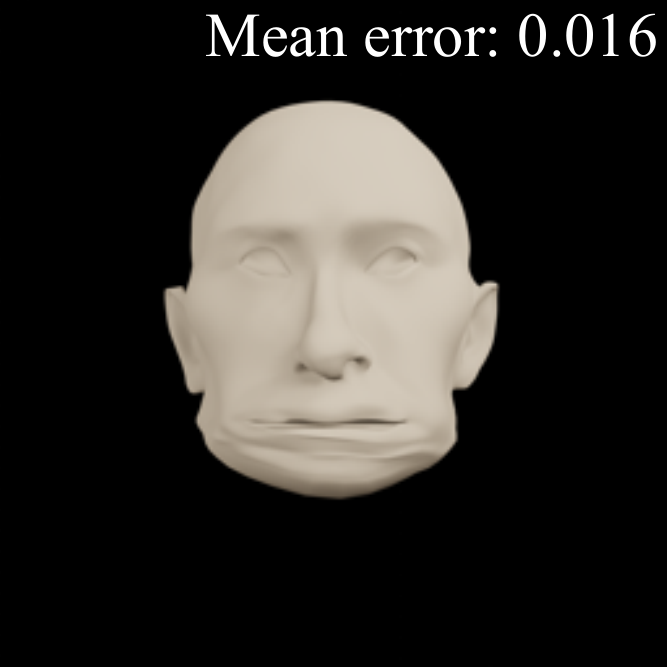}}\\ \\[4pt]
    
    Ground truth & Vertex array & Ours \\
    \end{tabular}
    \caption{Comparison to using vertex position array. Using our proposed representation and network architecture, the model provides faster convergence and better generalization. The reported error is the mean $l^2$ distance between corresponding vertices of the ground truth and the reconstruction. On average, the error for the vertex array case was 0.031 while ours achieved 0.017 on the test set.}
    \label{fig:comparison-fc}
\end{figure}

\Fig{comparison-fc} shows the reconstruction of an unseen 3D caricature shape using two different representations for the 3D caricature: vertex position array and our proposed continuous deformation function on the surface. Using vertex position array, we find it hard to train the model to generalize well to unseen caricatures. The design of the model using the vertex position array is inspired by \cite{guo20193d} and \cite{jiang2019disentangled}. Refer to the supplementary material for details on the network design and training of the model.

\begin{figure}
    \centering
    \small
    \renewcommand{\arraystretch}{0}
    
    \begin{tabular}{@{}c@{}c@{}c@{}c@{}}
    
    \raisebox{-.5\height}{\includegraphics[width=0.11\textwidth, height=0.11\textwidth]{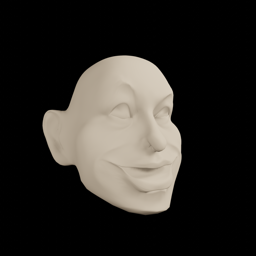}} &
    \raisebox{-.5\height}{\includegraphics[width=0.11\textwidth, height=0.11\textwidth]{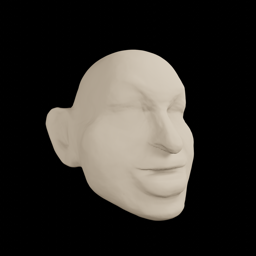}} &
    \raisebox{-.5\height}{\includegraphics[width=0.11\textwidth, height=0.11\textwidth]{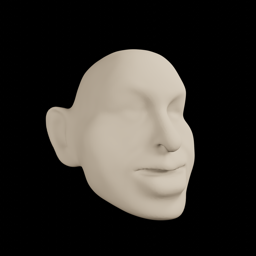}} &
    \raisebox{-.5\height}{\includegraphics[width=0.11\textwidth, height=0.11\textwidth]{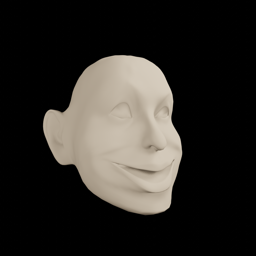}} \\

    \raisebox{-.5\height}{\includegraphics[width=0.11\textwidth, height=0.11\textwidth]{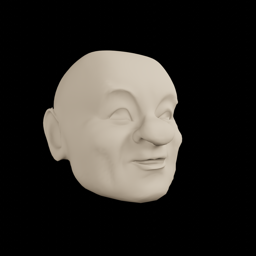}} &
    \raisebox{-.5\height}{\includegraphics[width=0.11\textwidth, height=0.11\textwidth]{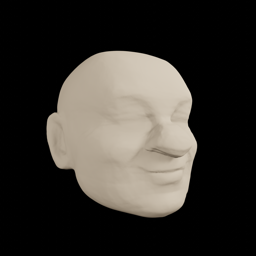}} &
    \raisebox{-.5\height}{\includegraphics[width=0.11\textwidth, height=0.11\textwidth]{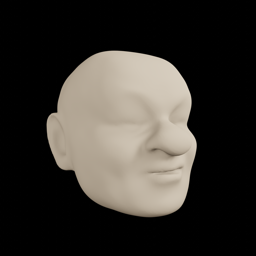}} &
    \raisebox{-.5\height}{\includegraphics[width=0.11\textwidth, height=0.11\textwidth]{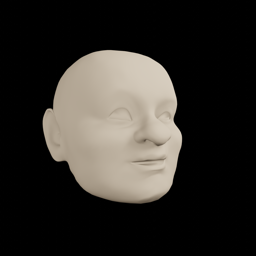}} \\ \\[4pt]
    
    Ground truth & DeepSDF & DIF-NET & Ours \\
    \end{tabular}
    \caption{Visual comparison to SDF-based auto-decoder methods. We show reconstructions of 3D shapes in the training set. While DeepSDF and DIF-Net reconstruct overall shapes reasonably, the reconstructions miss important details such as the shape of eyes or the expression around the mouth.}
    \label{fig:comparison-volume}
\end{figure}

\paragraph{SDF} \Fig{comparison-volume} compares visual results of training when SDF-based method is used to model 3D caricatures. We compare with DeepSDF \cite{park2019deepsdf} and DIF-Net \cite{deng2021deformed} using the authors' implementations. When we directly applied the auto-decoder framework for SDFs by converting all 3D caricatures into SDF, important details of the input were lost. Our method suffers less from the loss of details. One of the reasons is that we formulate our 3D caricature as a deformable model. Since we use a fixed template mesh that contains many crucial details, even when MLPs generate a low-frequency function, we readily obtain a reasonable 3D caricature shape due to existing details in the template.

\begin{figure*}[t]
\begin{tabular}{c c}
    \includegraphics[width=0.40\textwidth]{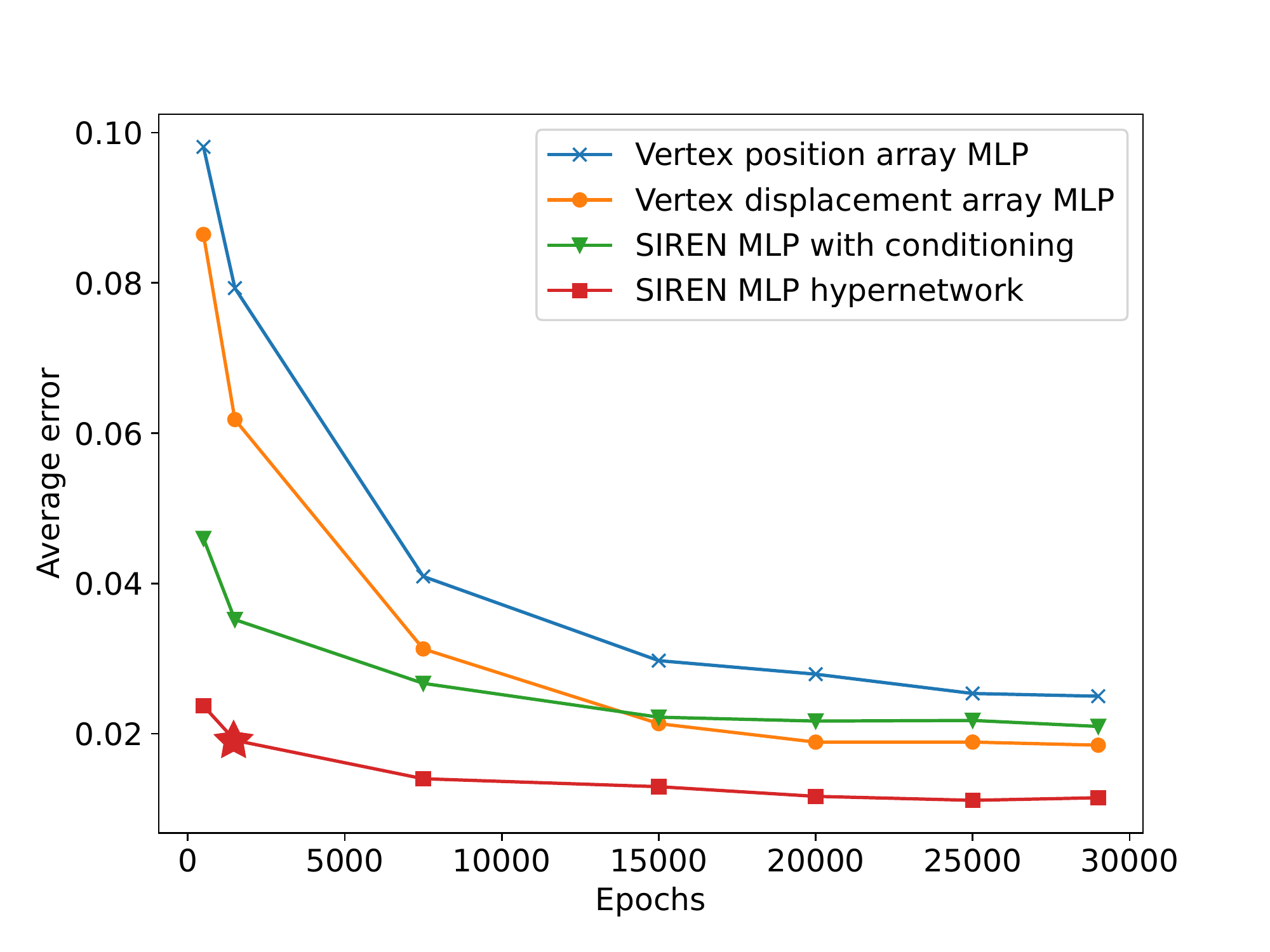} & \includegraphics[width=0.40\textwidth]{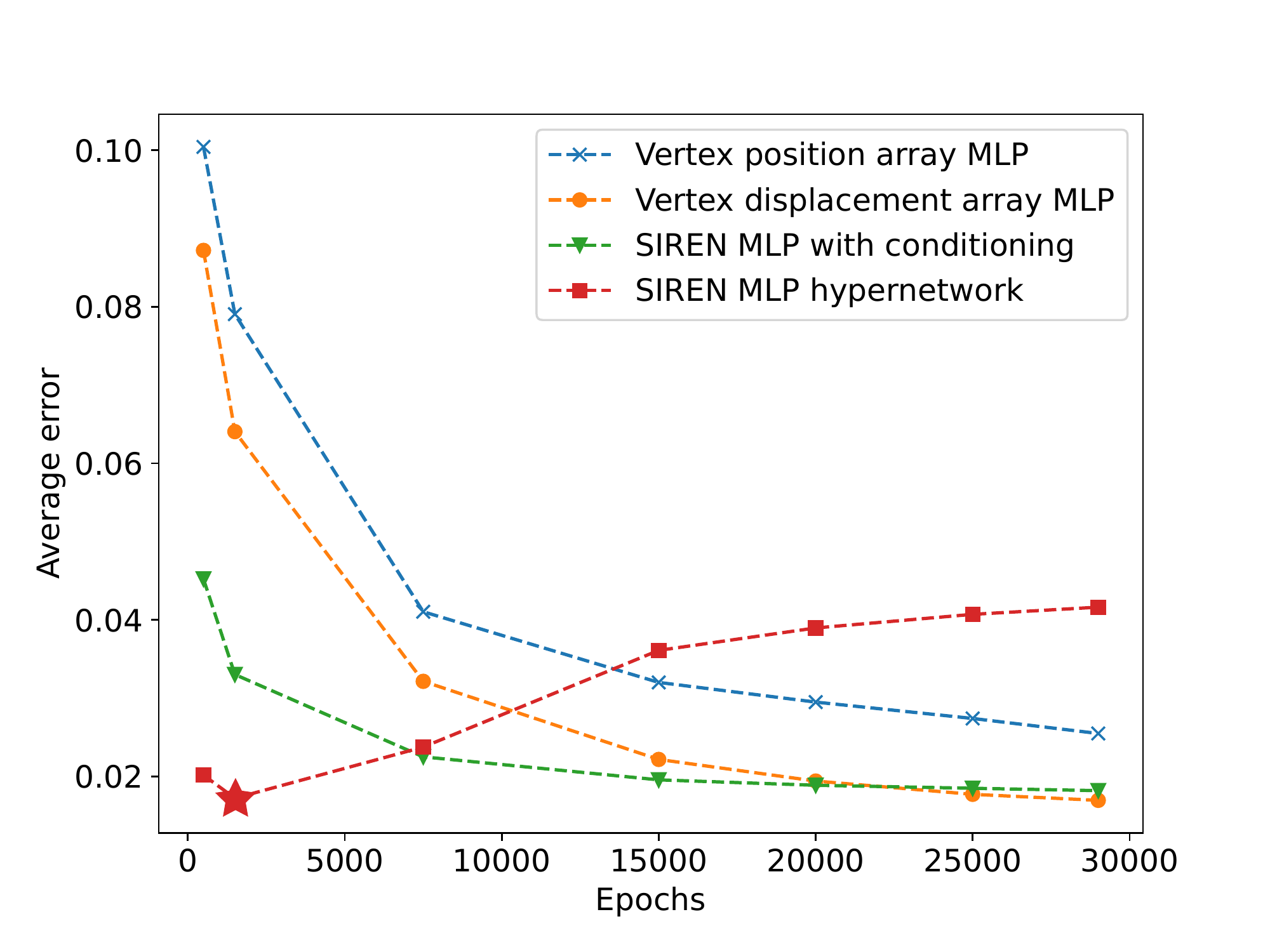} \\
\end{tabular}
\caption{Ablation study on the network architecture. We inspect error on the training set (left) and test set (right) of each model at different epochs. Star ($\star$) indicates early stopping at epoch 1500 determined by checking the validation error.}
\label{fig:ablation}
\end{figure*}

\subsection{Ablation study}
\label{sec:ablation-structure}

In this work, we designed the network architecture based on \textit{Deform-Net} of DIF-Net \cite{deng2021deformed}. We keep the network structure of \textit{Deform-Net} such as hypernetwork, while tweaking the training goal for the MLP to model template deformation. 
To inspect the effects of the components in our final model, we provide an ablation study of four different models.

\textit{Vertex position array MLP} is the baseline model discussed in \Sec{comparison}. The details for this model are in the supplementary document.

\textit{Vertex displacement array MLP} has the same network architecture as \textit{Vertex position array MLP}, but the model differs in the training goal so that the network's output is the per-vertex displacement from the template mesh.

\textit{SIREN MLP with conditioning} is a tweak of our final model. Instead of using hypernetwork, similarly to \cite{park2019deepsdf}, we introduce the shape variation by concatenating the latent code to the input template position and the features of the middle layer. 

\textit{SIREN MLP hypernetwork} is our proposed model in the main paper. We performed early-stopping at epoch 1500 using validation error to prevent over-fitting.

\Fig{ablation} provides a comparison of the average mean $l^2$ reconstruction error of different models at different epochs. Refer to the supplementary material for a visual comparison between the four models.

Comparing \textit{Vertex position array MLP} and \textit{Vertex displacement array MLP}, we observed lower training and test set error on the \textit{Vertex displacements array MLP}. The change in the representation from absolute position to deformation plays an important role in modeling 3D caricatures accurately.

Comparing \textit{SIREN MLP with conditioning} and \textit{SIREN MLP hypernetwork}, we observed lower training error on the \textit{SIREN MLP hypernetwork}. The hypernetwork provides higher network capacity. Furthermore, by comparing the error curve on the test set, we observed \textit{SIREN MLP hypernetwork} produces much faster convergence.

In conclusion, \textit{SIREN MLP hypernetwork} represents the complex 3D caricature data accurately by using the deformation representation. Moreover, the model converges fast by adopting the hypernetwork architecture.

\subsection{Continuous template deformation function}
\label{sec:continuous}

\begin{figure}
    \centering
    \begin{tabular}{c@{}c}
    \includegraphics[width=0.22\textwidth]{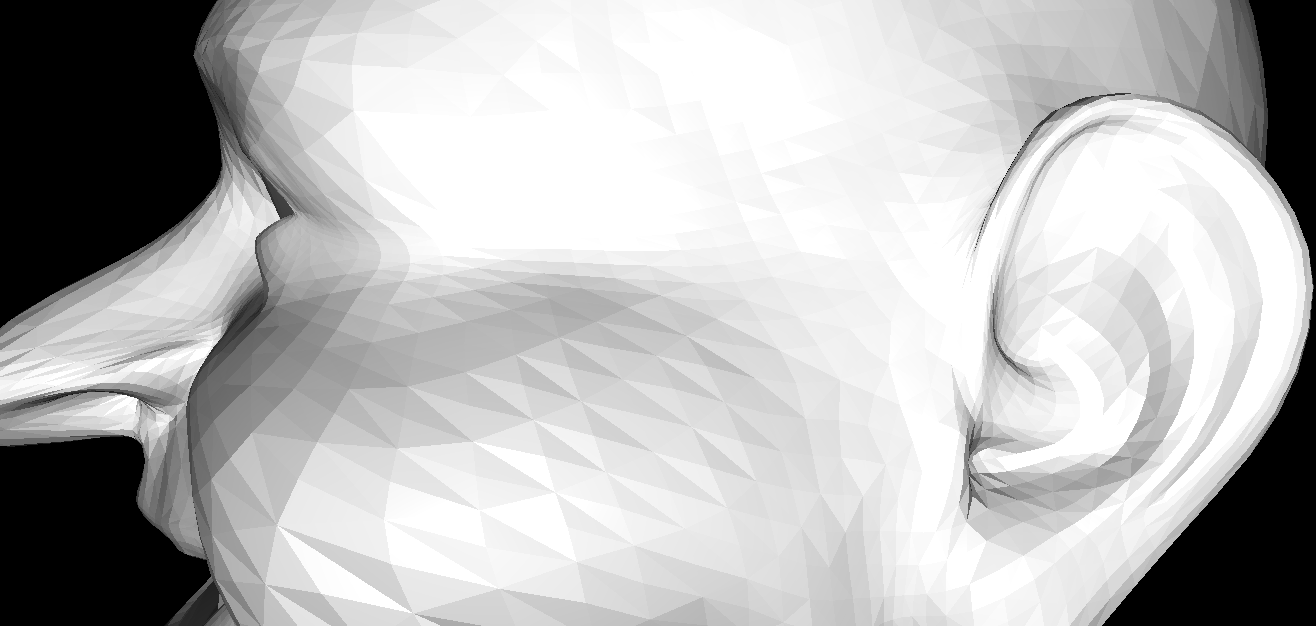} &
    \includegraphics[width=0.22\textwidth]{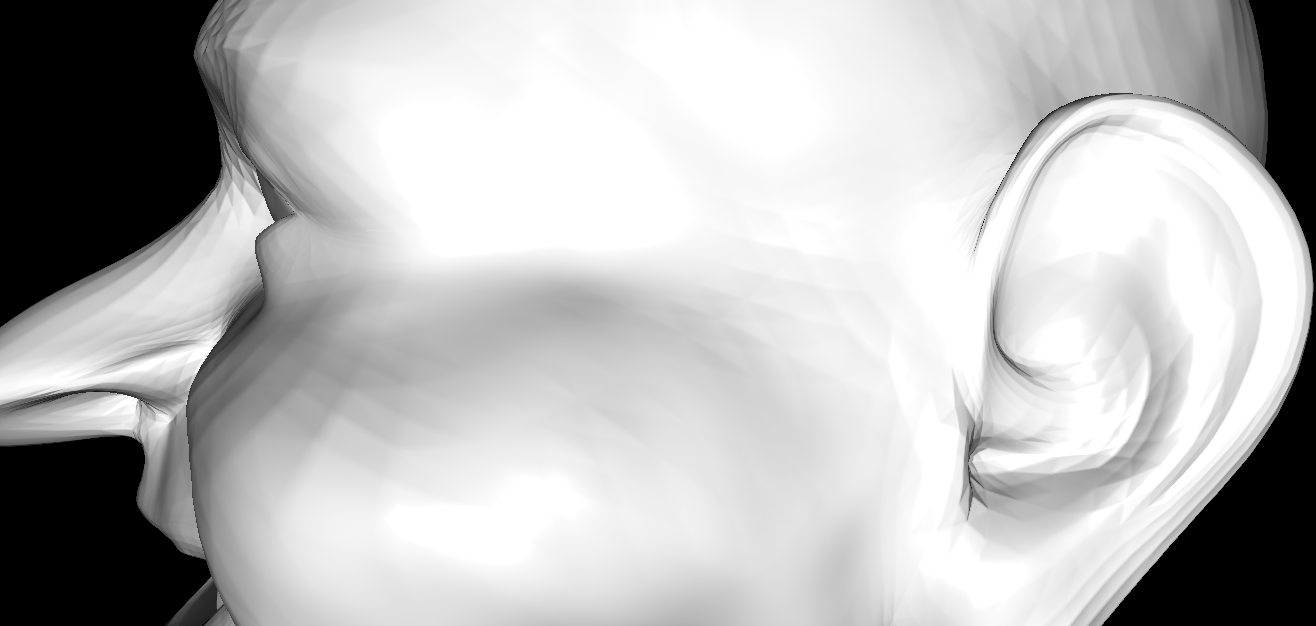} \\
    \end{tabular}
    \caption{Visualization of the continuous deformation function learned by the network. Left: A generated 3D caricature shape sampled with the original template mesh resolution. Right: The same caricature with a denser sampling on the template mesh. The continuous function modeled by the MLP is generally smooth, so the interiors of large triangles are filled with smooth shapes.}
    \label{fig:subdivision}
\end{figure}

Although we trained the network on a template mesh consisting of triangles, we can generate 3D caricatures of any resolution as the MLP models a continuous function defined on the template. We visualize a continuous deformation function generated from our network by showing a 3D caricature generated from an upsampled template mesh (\Fig{subdivision}). To upsample the template mesh, we subdivide each triangle into four triangles by splitting each edge,
as in Loop subdivision \cite{loop1987smooth}, but without moving initial vertex positions. We apply the subdivision operation three times to upsample the template mesh of 11,551 vertices to 739,183 vertices. The result shows that the generated deformation function favors spatially smooth deformation even though the training samples have been generated from the vertices on the mesh and the samples on each triangle. %

\section{Applications}

In this section, we demonstrate various applications using the learned latent space of our deep deformable 3D caricature model. In addition, we provide a supplementary video  that demonstrates our applications\footnote{\href{https://www.youtube.com/channel/UC3N03KqwKo_5gD4lFiSLMTg}{https://www.youtube.com/channel/UC3N03KqwKo\_5gD4lFiSLMTg}}. In the following, %
we refer to the deformable model $D$ as a function 
\begin{equation}
D(\hat{p}, z) = S(\hat{p}, H(z)) + \hat{p},
\end{equation}
where $S$ is SIREN MLP with parameters $H(z)$, $z$ is a latent code, and $\hat{p}$ is a point on the template surface.

\begin{figure}
    \centering
    \small
    \renewcommand{\arraystretch}{0}\begin{tabular}{@{}c@{}c@{}c@{}c@{}}
    
    \raisebox{-.5\height}{\includegraphics[width=0.22\columnwidth]{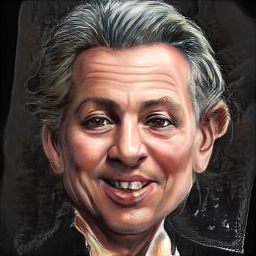}} &
    \raisebox{-.5\height}{\includegraphics[width=0.22\columnwidth]{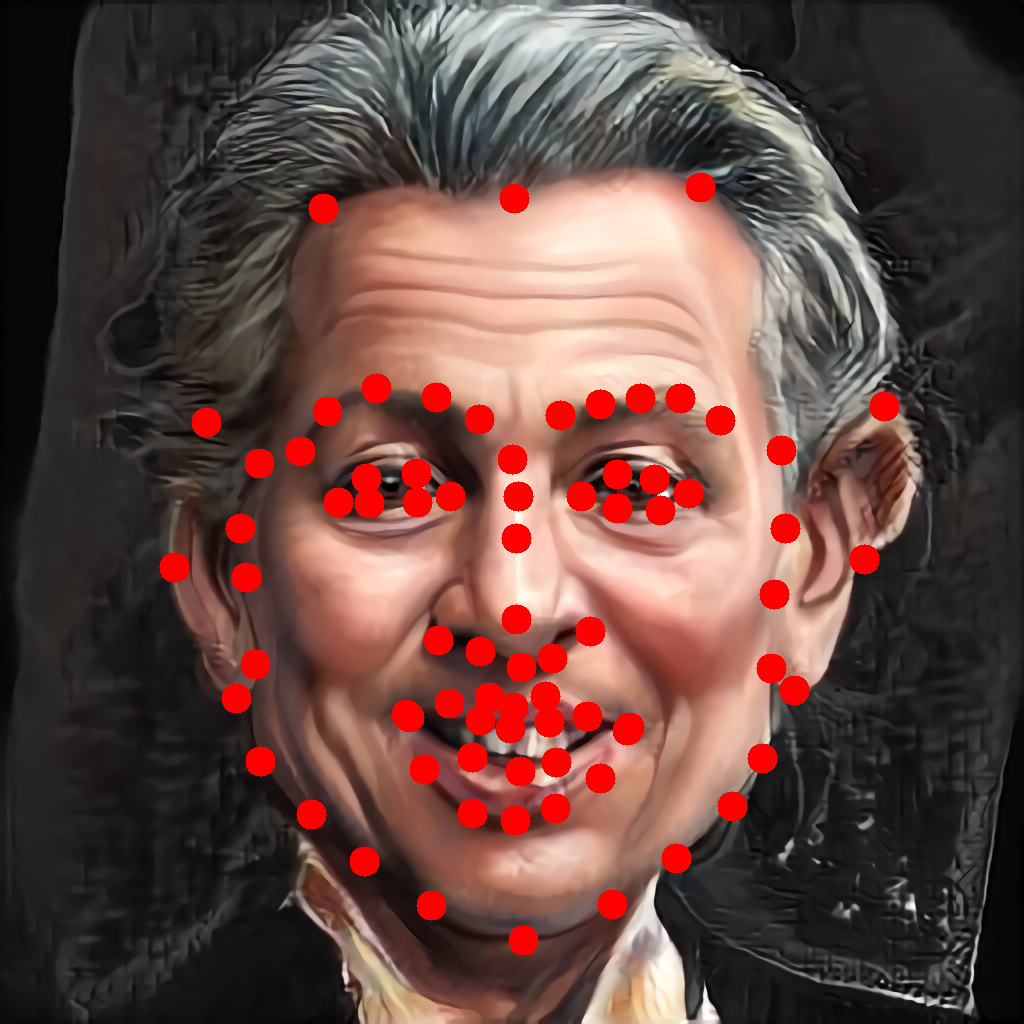}} &
    \raisebox{-.5\height}{\includegraphics[width=0.22\columnwidth]{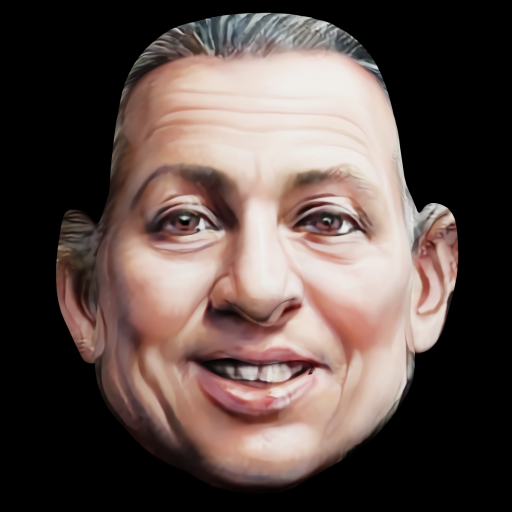}} &
    \raisebox{-.5\height}{\includegraphics[width=0.22\columnwidth]{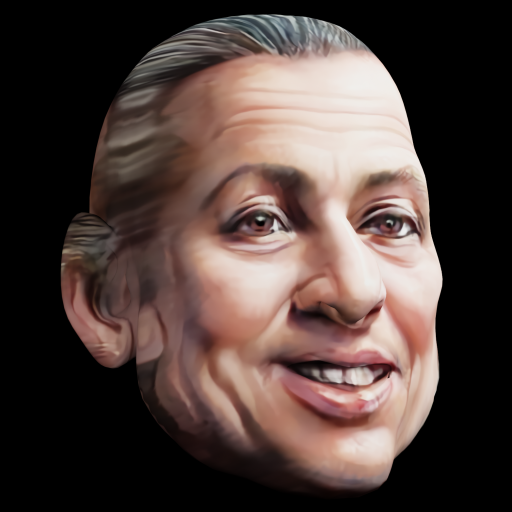}} \\
    
    \raisebox{-.5\height}{\includegraphics[width=0.22\columnwidth]{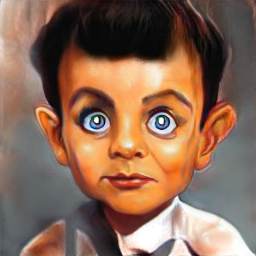}} &
    \raisebox{-.5\height}{\includegraphics[width=0.22\columnwidth]{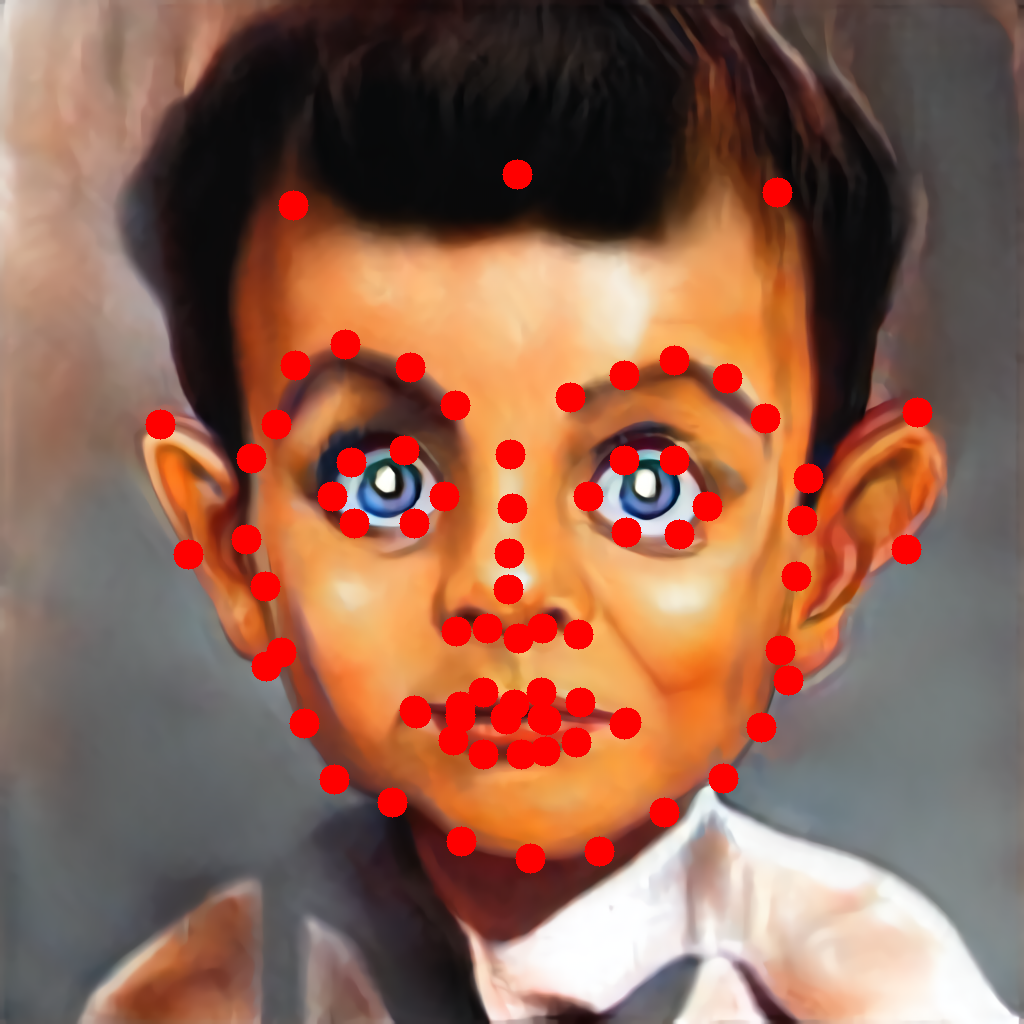}} &
    \raisebox{-.5\height}{\includegraphics[width=0.22\columnwidth]{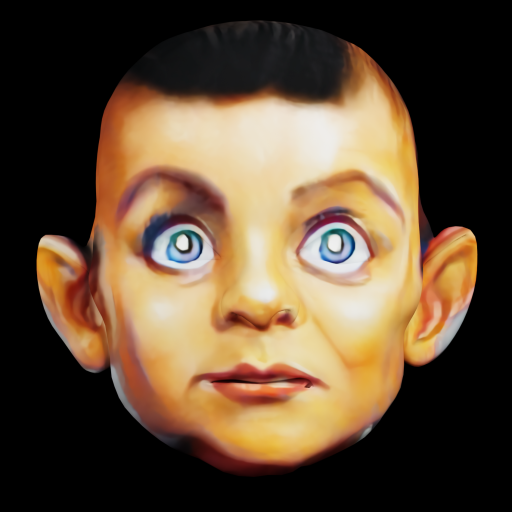}} &
    \raisebox{-.5\height}{\includegraphics[width=0.22\columnwidth]{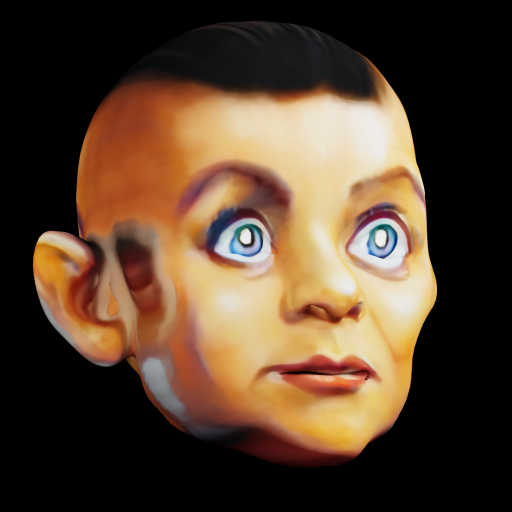}} \\
    
    \\[4pt]
    
    Input & Landmarks & \multicolumn{2}{c}{Reconstruction} \\
    
    \end{tabular}
    \caption{3D caricature reconstruction from 2D caricature landmarks. Using manually marked sparse landmarks on the input, we optimize the latent code for our model to fit the landmark constraints. The inputs are 2D caricatures generated by StyleCariGAN \cite{jang2021stylecarigan}.
    }
    \label{fig:reconstruction-scheme}
\end{figure}

\subsection{3D caricature reconstruction from 2D landmarks}
\label{sec:landmarkreconstruction}

In \Fig{reconstruction-scheme}, we demonstrate 3D caricature reconstruction from 2D landmarks.
Unlike previous work on 3D caricature reconstruction, our reconstruction is readily editable using the latent space. Alive Caricature \cite{wu2018alive} and 3DCaricShop \cite{qiu20213dcaricshop} almost perfectly fit the position constraints in the image by allowing free-form deformation of mesh vertices. Our approach reconstructs a 3D caricature by optimizing an editable latent code to reasonably fit the constraints in the input. For visual comparison with Alive Caricature and 3DCaricShop, refer to our project page\footnote{  \href{https://ycjungsubhuman.github.io/DeepDeformable3DCaricatures}{https://ycjungsubhuman.github.io/DeepDeformable3DCaricatures}}. 

Given 2D landmark positions $b$ and the corresponding 3D landmark vertex indices $l$, the fitting is done via an alternating optimization. In a similar fashion to \cite{wu2018alive}, we alternate between optimizing pose parameters $\Pi, R, t$ ($\Pi$ is an orthographic projection matrix, $R$ is a rotation matrix, $t$ is a translation vector) and the latent code for the shape $z$ to minimize the loss:
\begin{equation}
    L_{rec}(b, \hat{p}, z) =
    \dfrac{1}{B} \sum_i^{B} || \Pi R D(\hat{p}_{l(i)}, z) + t - b_i ||_2^2 + \dfrac{\lambda_{reg}}{M} || z ||_2^2,
\end{equation}
where $B$ is the number of 2D landmarks. and $l$ is landmark vertex indices. We iterate the pose and the shape optimizations four times. We provide more details on the optimization algorithm in the supplementary material.

\subsection{Sementic editing}

\begin{figure}
     \centering
     \small
     \begin{tabular}{@{\hspace{-4pt}}c@{\hspace{3pt}}c@{}c@{}c@{}c@{}c}
    
    (a) &
    \raisebox{-.5\height}{\includegraphics[width=0.09\textwidth, height=0.09\textwidth]{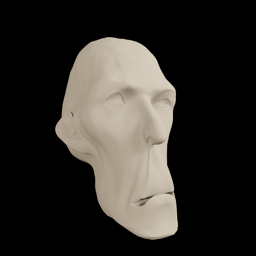}} &
    \raisebox{-.5\height}{\includegraphics[width=0.09\textwidth, height=0.09\textwidth]{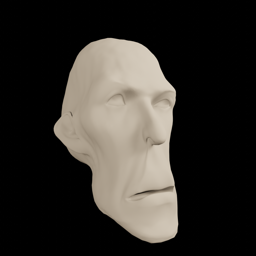}} &
    \raisebox{-.5\height}{\includegraphics[width=0.09\textwidth, height=0.09\textwidth]{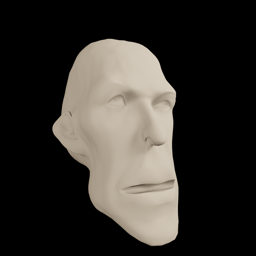}} &
    \raisebox{-.5\height}{\includegraphics[width=0.09\textwidth, height=0.09\textwidth]{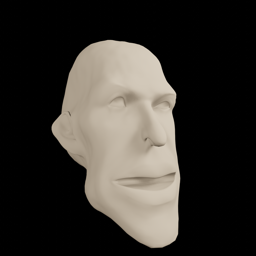}} &
    \raisebox{-.5\height}{\includegraphics[width=0.09\textwidth, height=0.09\textwidth]{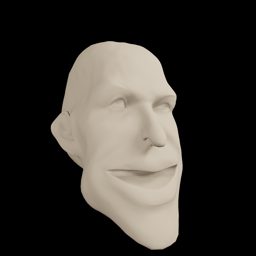}} \\
    
    (b) &
    \raisebox{-.5\height}{\includegraphics[width=0.09\textwidth, height=0.09\textwidth]{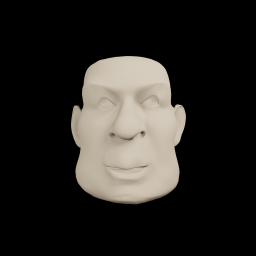}} &
    \raisebox{-.5\height}{\includegraphics[width=0.09\textwidth, height=0.09\textwidth]{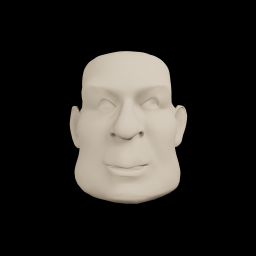}} &
    \raisebox{-.5\height}{\includegraphics[width=0.09\textwidth, height=0.09\textwidth]{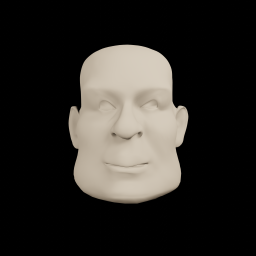}} &
    \raisebox{-.5\height}{\includegraphics[width=0.09\textwidth, height=0.09\textwidth]{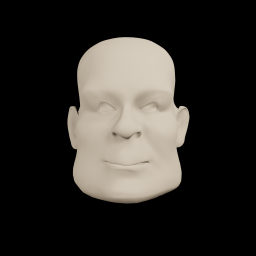}} &
    \raisebox{-.5\height}{\includegraphics[width=0.09\textwidth, height=0.09\textwidth]{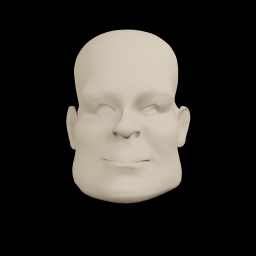}} \\
    
    \vspace{4pt}
    (c) &
    \raisebox{-.5\height}{\includegraphics[width=0.09\textwidth, height=0.09\textwidth]{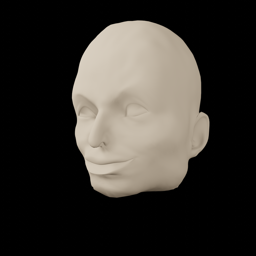}} &
    \raisebox{-.5\height}{\includegraphics[width=0.09\textwidth, height=0.09\textwidth]{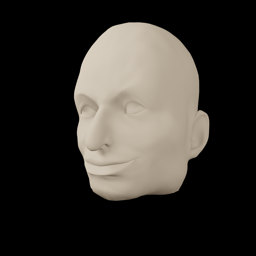}} &
    \raisebox{-.5\height}{\includegraphics[width=0.09\textwidth, height=0.09\textwidth]{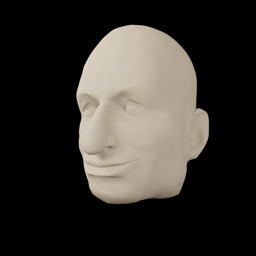}} &
    \raisebox{-.5\height}{\includegraphics[width=0.09\textwidth, height=0.09\textwidth]{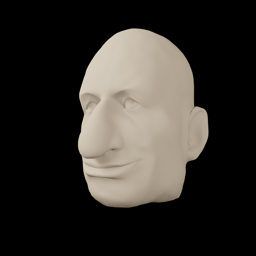}} &
    \raisebox{-.5\height}{\includegraphics[width=0.09\textwidth, height=0.09\textwidth]{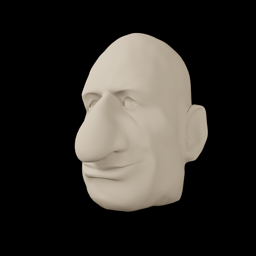}} \\
    
     & \multicolumn{2}{c}{$(-)$} & Original & \multicolumn{2}{c}{$(+)$} \\
    \end{tabular}
    
     \caption{Semantic editing using InterFaceGAN technique. (a) Editing on the label \textit{Smile}. (b) \textit{Large forehead}. (c) \textit{Big nose}. $(+)$ denotes adding the editing vector that corresponds to the direction of the label on each row. $(-)$ denotes subtracting the vector.}
     \label{fig:editing-semantic}
\end{figure}

We can apply semantic editing by manipulating the latent code of an input 3D face in a similar fashion to 2D caricature editing demonstrated in StyleCariGAN \cite{jang2021stylecarigan}. The WebCariA dataset \cite{ji2020unsupervised} provides a set of semantic labels, e.g., whether a caricature is smiling or not, for each caricature in the training set. Using InterFaceGAN technique \cite{shen2020interfacegan} to our latent space, we obtain the direction for each attribute in the WebCariA dataset. \Fig{editing-semantic} demonstrates the semantic editing results on the reconstructions of the test split of the 3DCaricShop dataset.

\begin{figure}[t]
    \centering
    \small
    \renewcommand{\arraystretch}{0}\begin{tabular}{c@{}c@{}c@{}c@{}c}
    
    \raisebox{-.5\height}{\includegraphics[width=0.09\textwidth, height=0.09\textwidth]{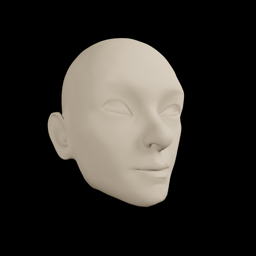}} &
    \raisebox{-.5\height}{\includegraphics[width=0.09\textwidth, height=0.09\textwidth]{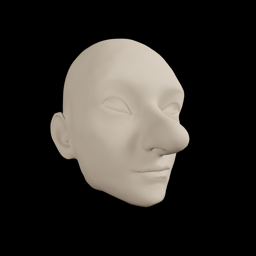}} &
    \raisebox{-.5\height}{\includegraphics[width=0.09\textwidth, height=0.09\textwidth]{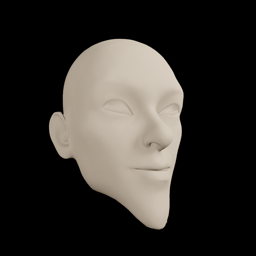}} &
    \raisebox{-.5\height}{\includegraphics[width=0.09\textwidth, height=0.09\textwidth]{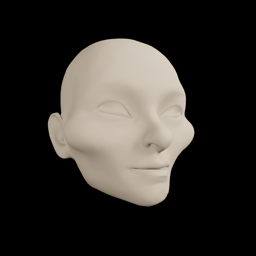}} &
    \raisebox{-.5\height}{\includegraphics[width=0.09\textwidth, height=0.09\textwidth]{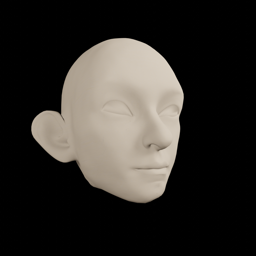}}   \\
    
    \raisebox{-.5\height}{\includegraphics[width=0.09\textwidth, height=0.09\textwidth]{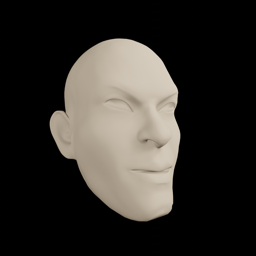}} &
    \raisebox{-.5\height}{\includegraphics[width=0.09\textwidth, height=0.09\textwidth]{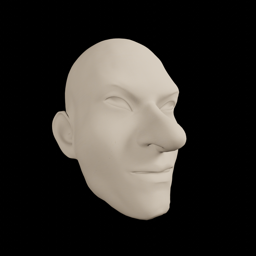}} &
    \raisebox{-.5\height}{\includegraphics[width=0.09\textwidth, height=0.09\textwidth]{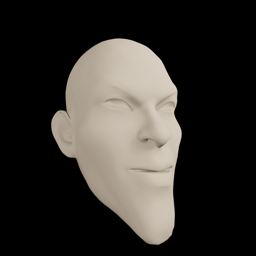}} &
    \raisebox{-.5\height}{\includegraphics[width=0.09\textwidth, height=0.09\textwidth]{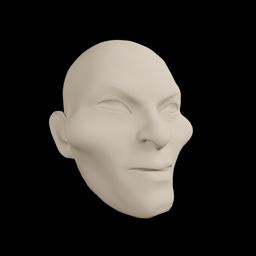}} &
    \raisebox{-.5\height}{\includegraphics[width=0.09\textwidth, height=0.09\textwidth]{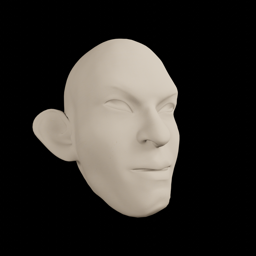}}   \\
    
    \raisebox{-.5\height}{\includegraphics[width=0.09\textwidth, height=0.09\textwidth]{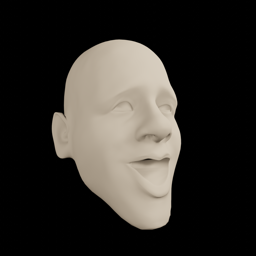}} &
    \raisebox{-.5\height}{\includegraphics[width=0.09\textwidth, height=0.09\textwidth]{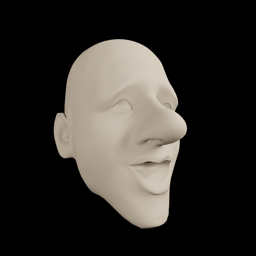}} &
    \raisebox{-.5\height}{\includegraphics[width=0.09\textwidth, height=0.09\textwidth]{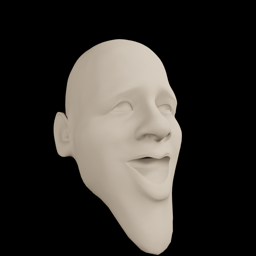}} &
    \raisebox{-.5\height}{\includegraphics[width=0.09\textwidth, height=0.09\textwidth]{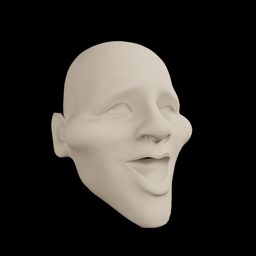}} &
    \raisebox{-.5\height}{\includegraphics[width=0.09\textwidth, height=0.09\textwidth]{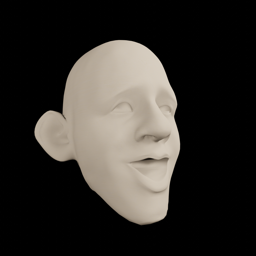}} \\ \\[4pt]
    
    Original & Nose & Chin & Cheek & Ears \\
    \end{tabular}
    \caption{Point-handle-based editing.
    Nose: Pick a point on the nose tip. Move the point to the front. Chin: Pick a point on the bottom of the chin. Move the point to the bottom. Cheek: Pick a point at each side of the cheek. Stretch the two points sideways. Ears: Pick a point at each side of the ears. Stretch the two points. }
    \label{fig:editing-handle}
\end{figure}

\subsection{Point-handle-based editing}
\label{sec:pointediting}

Using the learned latent space, we can apply point-handle-based editing to a 3D caricature shape. \Fig{editing-handle} shows editing on the 3D caricature reconstructions from 2D caricatures. Even though the input for editing is extremely sparse, e.g., one or two points, it produces plausible deformation to complete local editing. Given a latent code $z_0$ for the initial 3D shape to be edited, a list of vertex indices $h$ for the handles, and the 3D displacement vectors $\delta$ for each handle, we solve an optimization problem that minimizes
\begin{equation}
    L_{edit}(\delta, \hat{p}, z) = \dfrac{\lambda_{con}}{H} \sum_i^{H} || p_{h(i)} + \delta_{i} - D(\hat{p}_{h(i)}, z) ||_2^2 + \dfrac{\lambda_{pre}}{M} || z_0 - z ||_1,
\end{equation}
where $H$ is the number of handles, $p$ is the initial shape from $z_0$, the weight for the constraint term $\lambda_{con} = 3.0 \times 10^{3}$, and the weight for term that enforces the preservation of initial shape $\lambda_{pre} = 1.0 \times 10^{5}$.

Our handle-based deformation can be compared to other traditional handle-based surface deformation technique such as as-rigid-as-possible (ARAP) mesh deformation \cite{sorkine2007as}. Since ARAP energy is insensitive to global translation by design, the optimal deformation given a single point handle displacement is naturally a global translation without any local deformation. To create local deformation using a single point in ARAP, manual selection of the region of interest (ROI) is required. Our method produces natural local deformation even with a single point constraint. Also, expansion or exaggeration, which we believe is crucial for caricatures, is not easily allowed in ARAP deformation. Our method naturally favors local expansion using the learned latent space. We provide visual comparison with ARAP in the supplementary material.

\subsection {Automatic 3D caricature creation} 

\begin{figure}
    \centering
    \small
    \renewcommand{\arraystretch}{0}
    
    \begin{tabular}{@{}c@{}c@{}c@{}}
    
    \includegraphics[width=0.11\textwidth, height=0.11\textwidth]{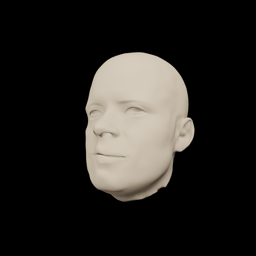} &
    \includegraphics[width=0.11\textwidth, height=0.11\textwidth]{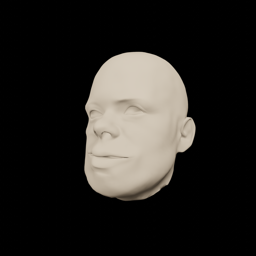} &
    \includegraphics[width=0.11\textwidth, height=0.11\textwidth]{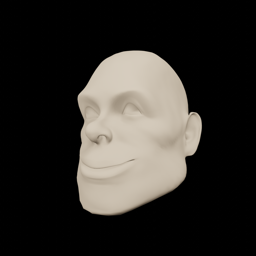}  \\
    
    \includegraphics[width=0.11\textwidth, height=0.11\textwidth]{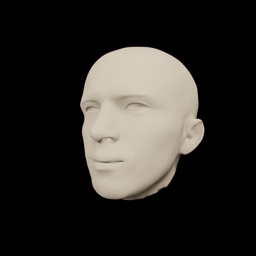} &
    \includegraphics[width=0.11\textwidth, height=0.11\textwidth]{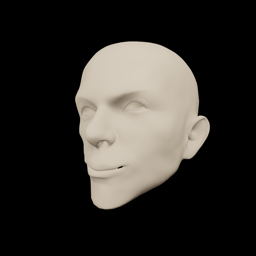} &
    \includegraphics[width=0.11\textwidth, height=0.11\textwidth]{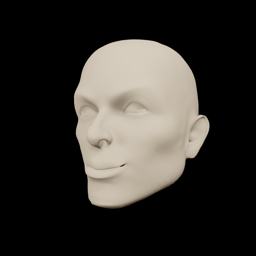}\\  \\[4pt]

    Original & \multicolumn{2}{c}{Caricature} \\
    \end{tabular}
     \caption{Automatically created caricatures of varying degrees. Given a regular face, we apply semantic editing towards caricatures. The third column is edited three times further towards caricatures compared to the second column.} %
     \label{fig:automatic-caricature}
\end{figure}

\Fig{automatic-caricature} demonstrates automatic caricature creation for a regular 3D face through semantic editing. By training our model with both 3D caricatures and regular 3D faces, we obtain a model that spans both the 3D caricatures and regular faces. By calculating the editing direction from the class of regular faces to caricature faces using InterFaceGAN \cite{shen2020interfacegan}, we enable the editing operation to be applied to the latent code of a regular face. 

To enable this editing, we train our model using both 3DCaricShop \cite{qiu20213dcaricshop} and FaceWarehouse \cite{cao2013facewarehouse} that provide 3D caricature examples and regular 3D face examples, respectively. The model was trained using the training set of 3DCaricShop and all neutral faces in FaceWarehouse. Automatic caricature generation is done starting from a neutral face in the FaceWarehouse dataset. We describe details on FaceWarehouse dataset processing in the supplementary material.

\begin{figure}
    \centering
    \small
    \renewcommand{\arraystretch}{0}\begin{tabular}{@{}c@{}c@{}c@{}c@{}c}
    
    \raisebox{-.5\height}{\includegraphics[width=0.19\columnwidth]{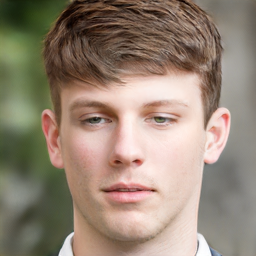}} &
    \raisebox{-.5\height}{\includegraphics[width=0.19\columnwidth]{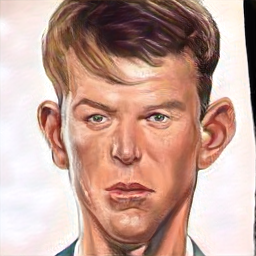}} &
    \raisebox{-.5\height}{\includegraphics[width=0.19\columnwidth]{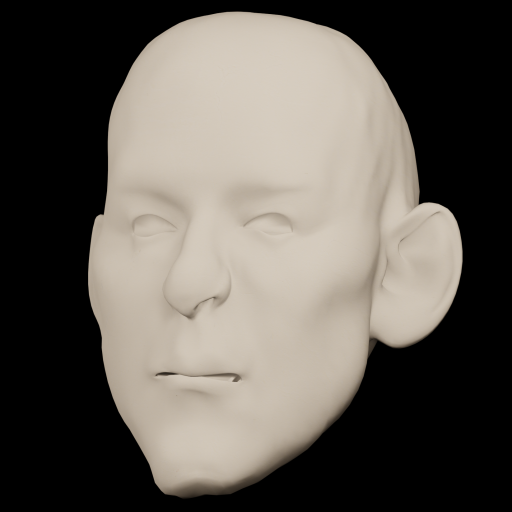}} &
    \raisebox{-.5\height}{\includegraphics[width=0.19\columnwidth]{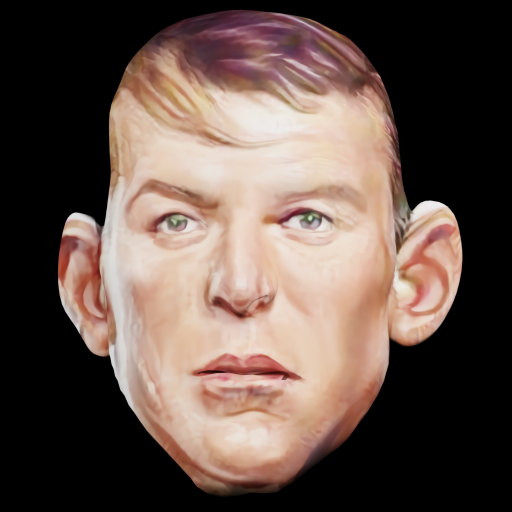}} &
    \raisebox{-.5\height}{\includegraphics[width=0.19\columnwidth]{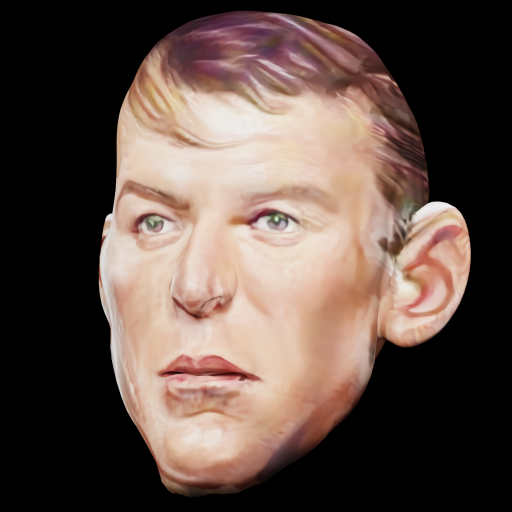}}
    \\
    
    \raisebox{-.5\height}{\includegraphics[width=0.19\columnwidth]{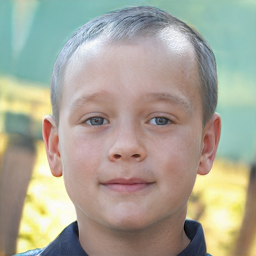}} &
    \raisebox{-.5\height}{\includegraphics[width=0.19\columnwidth]{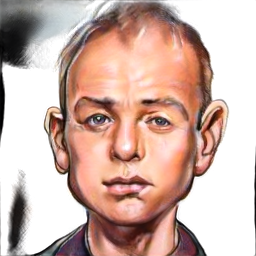}} &
    \raisebox{-.5\height}{\includegraphics[width=0.19\columnwidth]{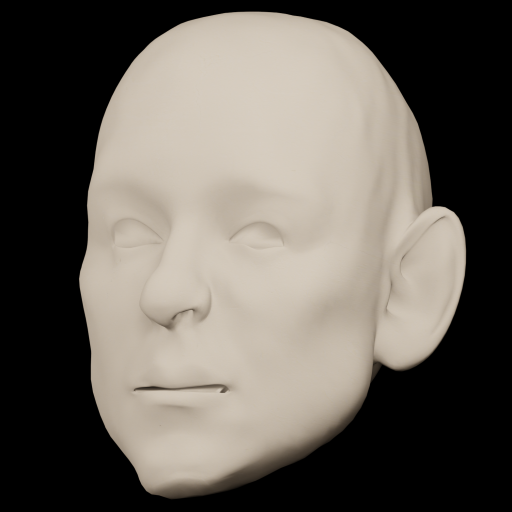}} &
    \raisebox{-.5\height}{\includegraphics[width=0.19\columnwidth]{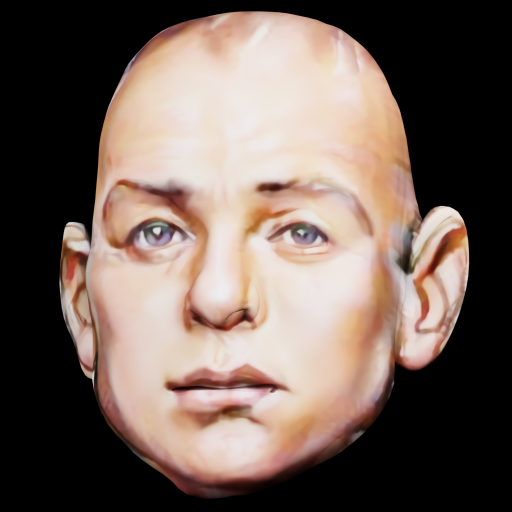}} &
    \raisebox{-.5\height}{\includegraphics[width=0.19\columnwidth]{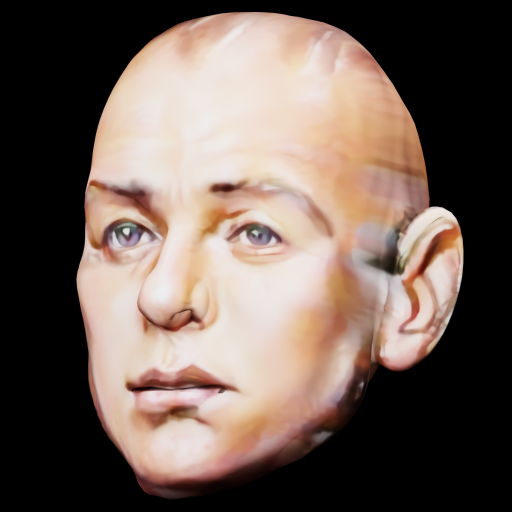}} \\ \\[4pt]
    
    Input & Caricature & 3D & \multicolumn{2}{c}{Textured} \\
    
    \end{tabular}
    \caption{Automatic 3D caricature creation via 3D caricature reconstruction on the results of StyleCariGAN. The inputs are photos generated by StyleGAN \cite{karras2019style}. This application also works for real-world photo inputs.
    }
    \label{fig:stylecarigan}
\end{figure}

Another approach for automatic caricature creation is to utilize a 2D caricature generator. Given an image of a subject, we generate a 2D caricature using StyleCariGAN \cite{jang2021stylecarigan}. Using manually marked landmarks on the generated caricature, we can create an editable automatic 3D caricature (\Fig{stylecarigan}). %

\subsection {Running time} 
We implemented and tested our model using PyTorch \cite{NEURIPS2019_9015}, and run it on a PC with an AMD Ryzen 9 3950X and an NVIDIA Titan Xp GPU. Generating a 3D shape with 11,551 vertices takes 17 ms, and a 3D shape with 739,183 vertices takes 987~ms (\Sec{continuous}). The alternating optimization for our 3D caricature reconstruction from 2D landmarks takes approximately 500~ms~(\Sec{landmarkreconstruction}). The optimization for our point-handle-based editing took approximately 20~ms (\Sec{pointediting}).

\section{Conclusion}

To build a useful editing space for 3D caricatures, we proposed a deformable 3D caricature framework. By adopting MLPs to model a continuous function of template surface deformation, we model the complex variations of 3D caricatures effectively. Once the model has been learned, it can be readily used for other tasks by editing latent codes or optimizing the latent codes to meet various constraints.

An interesting implication of our method is that each caricature is modeled as a continuous and differentiable function. This setting provides easy access to differential properties of deformation without using finite element methods. For example, by inspecting MLP parameters, we may easily inspect the smoothness of deformation function as demonstrated for SDFs in \cite{liu2022learning}.

Our network may not completely restore high-frequency details originally present in ground-truth 3D caricatures (See Figs.~\ref{fig:comparison-fc} and~\ref{fig:comparison-volume}). We also observed self-intersection around the chin from some reconstruction results, where training data often contain artifacts. Our point-handle-based editing may result in self-intersection given some input since we do not put any explicit protection from self-intersection. Improving the details and preventing self-intersections are left as future work. An explicit guarantee for identity preservation in editing would also be an interesting future work.

We believe our method can be applied to other domains where the data to be modeled are surfaces in dense correspondence. Our method assumes the template mesh does not have self-intersections. When the assumption does not hold, a different position encoding scheme would be required to differentiate two different points on the template surface sharing the same coordinates.

\begin{acks}
We would like to thank the anonymous reviewers for their constructive comments. 
This work was supported by 
IITP grants 
(SW Star Lab, 2015-0-00174;
Collaborative Research Project with MSRA, IITP-2020-0-01649;
AI Innovation Hub, 2021-0-02068; 
AI Graduate School Program (POSTECH), 2019-0-01906)
and
KOCCA grant (R2021040136) 
from Korea government (MSIT and MCST).
\end{acks}

\bibliographystyle{ACM-Reference-Format}
\bibliography{main}


\begin{thebibliography}{42}


\ifx \showCODEN    \undefined \def \showCODEN     #1{\unskip}     \fi
\ifx \showDOI      \undefined \def \showDOI       #1{#1}\fi
\ifx \showISBNx    \undefined \def \showISBNx     #1{\unskip}     \fi
\ifx \showISBNxiii \undefined \def \showISBNxiii  #1{\unskip}     \fi
\ifx \showISSN     \undefined \def \showISSN      #1{\unskip}     \fi
\ifx \showLCCN     \undefined \def \showLCCN      #1{\unskip}     \fi
\ifx \shownote     \undefined \def \shownote      #1{#1}          \fi
\ifx \showarticletitle \undefined \def \showarticletitle #1{#1}   \fi
\ifx \showURL      \undefined \def \showURL       {\relax}        \fi
\providecommand\bibfield[2]{#2}
\providecommand\bibinfo[2]{#2}
\providecommand\natexlab[1]{#1}
\providecommand\showeprint[2][]{arXiv:#2}

\bibitem[Blanz et~al\mbox{.}(1999)]%
        {blanz1999morphable}
\bibfield{author}{\bibinfo{person}{Volker Blanz}, \bibinfo{person}{Thomas
  Vetter}, {et~al\mbox{.}}} \bibinfo{year}{1999}\natexlab{}.
\newblock \showarticletitle{A morphable model for the synthesis of 3D faces}.
  In \bibinfo{booktitle}{\emph{Proc. SIGGRAPH}}.
\newblock


\bibitem[Cai et~al\mbox{.}(2021)]%
        {cai2021landmark}
\bibfield{author}{\bibinfo{person}{Hongrui Cai}, \bibinfo{person}{Yudong Guo},
  \bibinfo{person}{Zhuang Peng}, {and} \bibinfo{person}{Juyong Zhang}.}
  \bibinfo{year}{2021}\natexlab{}.
\newblock \showarticletitle{Landmark detection and 3D face reconstruction for
  caricature using a nonlinear parametric model}.
\newblock \bibinfo{journal}{\emph{Graphical Models}}  \bibinfo{volume}{115}
  (\bibinfo{year}{2021}), \bibinfo{pages}{101103}.
\newblock


\bibitem[Cao et~al\mbox{.}(2013)]%
        {cao2013facewarehouse}
\bibfield{author}{\bibinfo{person}{Chen Cao}, \bibinfo{person}{Yanlin Weng},
  \bibinfo{person}{Shun Zhou}, \bibinfo{person}{Yiying Tong}, {and}
  \bibinfo{person}{Kun Zhou}.} \bibinfo{year}{2013}\natexlab{}.
\newblock \showarticletitle{Facewarehouse: A 3d facial expression database for
  visual computing}.
\newblock \bibinfo{journal}{\emph{IEEE Trans. Vis. Comput. Graph.}}
  \bibinfo{volume}{20}, \bibinfo{number}{3} (\bibinfo{year}{2013}),
  \bibinfo{pages}{413--425}.
\newblock


\bibitem[Chan et~al\mbox{.}(2021)]%
        {chan2021pi}
\bibfield{author}{\bibinfo{person}{Eric~R Chan}, \bibinfo{person}{Marco
  Monteiro}, \bibinfo{person}{Petr Kellnhofer}, \bibinfo{person}{Jiajun Wu},
  {and} \bibinfo{person}{Gordon Wetzstein}.} \bibinfo{year}{2021}\natexlab{}.
\newblock \showarticletitle{pi-gan: Periodic implicit generative adversarial
  networks for 3d-aware image synthesis}. In \bibinfo{booktitle}{\emph{Proc.
  CVPR}}.
\newblock


\bibitem[Cheng et~al\mbox{.}(2019)]%
        {cheng2019meshgan}
\bibfield{author}{\bibinfo{person}{Shiyang Cheng}, \bibinfo{person}{Michael
  Bronstein}, \bibinfo{person}{Yuxiang Zhou}, \bibinfo{person}{Irene Kotsia},
  \bibinfo{person}{Maja Pantic}, {and} \bibinfo{person}{Stefanos Zafeiriou}.}
  \bibinfo{year}{2019}\natexlab{}.
\newblock \showarticletitle{Meshgan: Non-linear 3d morphable models of faces}.
\newblock \bibinfo{journal}{\emph{arXiv preprint arXiv:1903.10384}}
  (\bibinfo{year}{2019}).
\newblock


\bibitem[Deng et~al\mbox{.}(2021)]%
        {deng2021deformed}
\bibfield{author}{\bibinfo{person}{Yu Deng}, \bibinfo{person}{Jiaolong Yang},
  {and} \bibinfo{person}{Xin Tong}.} \bibinfo{year}{2021}\natexlab{}.
\newblock \showarticletitle{Deformed implicit field: Modeling 3d shapes with
  learned dense correspondence}. In \bibinfo{booktitle}{\emph{Proc. CVPR}}.
\newblock


\bibitem[Deng et~al\mbox{.}(2019)]%
        {deng2019accurate}
\bibfield{author}{\bibinfo{person}{Yu Deng}, \bibinfo{person}{Jiaolong Yang},
  \bibinfo{person}{Sicheng Xu}, \bibinfo{person}{Dong Chen},
  \bibinfo{person}{Yunde Jia}, {and} \bibinfo{person}{Xin Tong}.}
  \bibinfo{year}{2019}\natexlab{}.
\newblock \showarticletitle{Accurate 3D Face Reconstruction with
  Weakly-Supervised Learning: From Single Image to Image Set}. In
  \bibinfo{booktitle}{\emph{Proc. CVPRW}}.
\newblock


\bibitem[Feng et~al\mbox{.}(2018)]%
        {feng2018joint}
\bibfield{author}{\bibinfo{person}{Yao Feng}, \bibinfo{person}{Fan Wu},
  \bibinfo{person}{Xiaohu Shao}, \bibinfo{person}{Yanfeng Wang}, {and}
  \bibinfo{person}{Xi Zhou}.} \bibinfo{year}{2018}\natexlab{}.
\newblock \showarticletitle{Joint 3d face reconstruction and dense alignment
  with position map regression network}. In \bibinfo{booktitle}{\emph{Proc.
  ECCV}}.
\newblock


\bibitem[Garrido et~al\mbox{.}(2016)]%
        {garrido2016reconstruction}
\bibfield{author}{\bibinfo{person}{Pablo Garrido}, \bibinfo{person}{Michael
  Zollh{\"o}fer}, \bibinfo{person}{Dan Casas}, \bibinfo{person}{Levi
  Valgaerts}, \bibinfo{person}{Kiran Varanasi}, \bibinfo{person}{Patrick
  P{\'e}rez}, {and} \bibinfo{person}{Christian Theobalt}.}
  \bibinfo{year}{2016}\natexlab{}.
\newblock \showarticletitle{Reconstruction of personalized 3D face rigs from
  monocular video}.
\newblock \bibinfo{journal}{\emph{ACM Transactions on Graphics}}
  \bibinfo{volume}{35}, \bibinfo{number}{3} (\bibinfo{year}{2016}),
  \bibinfo{pages}{28}.
\newblock


\bibitem[Ghafourzadeh et~al\mbox{.}(2020)]%
        {partbased}
\bibfield{author}{\bibinfo{person}{Donya Ghafourzadeh}, \bibinfo{person}{Cyrus
  Rahgoshay}, \bibinfo{person}{Sahel Fallahdoust}, \bibinfo{person}{Andre
  Beauchamp}, \bibinfo{person}{Adeline Aubame}, \bibinfo{person}{Tiberiu Popa},
  {and} \bibinfo{person}{Eric Paquette}.} \bibinfo{year}{2020}\natexlab{}.
\newblock \showarticletitle{Part-Based 3D Face Morphable Model with
  Anthropometric Local Control}. In \bibinfo{booktitle}{\emph{Proc. GI}}.
\newblock


\bibitem[Gropp et~al\mbox{.}(2020)]%
        {gropp2020implicit}
\bibfield{author}{\bibinfo{person}{Amos Gropp}, \bibinfo{person}{Lior Yariv},
  \bibinfo{person}{Niv Haim}, \bibinfo{person}{Matan Atzmon}, {and}
  \bibinfo{person}{Yaron Lipman}.} \bibinfo{year}{2020}\natexlab{}.
\newblock \showarticletitle{Implicit Geometric Regularization for Learning
  Shapes}. In \bibinfo{booktitle}{\emph{Proc. MLR}}.
\newblock


\bibitem[Groueix et~al\mbox{.}(2018)]%
        {groueix2018papier}
\bibfield{author}{\bibinfo{person}{Thibault Groueix}, \bibinfo{person}{Matthew
  Fisher}, \bibinfo{person}{Vladimir~G Kim}, \bibinfo{person}{Bryan~C Russell},
  {and} \bibinfo{person}{Mathieu Aubry}.} \bibinfo{year}{2018}\natexlab{}.
\newblock \showarticletitle{A papier-m{\^a}ch{\'e} approach to learning 3d
  surface generation}. In \bibinfo{booktitle}{\emph{Proc. CVPR}}.
\newblock


\bibitem[Guo et~al\mbox{.}(2019)]%
        {guo20193d}
\bibfield{author}{\bibinfo{person}{Yudong Guo}, \bibinfo{person}{Luo Jiang},
  \bibinfo{person}{Lin Cai}, {and} \bibinfo{person}{Juyong Zhang}.}
  \bibinfo{year}{2019}\natexlab{}.
\newblock \showarticletitle{3D Magic Mirror: Automatic Video to 3D Caricature
  Translation}.
\newblock \bibinfo{journal}{\emph{arXiv preprint arXiv:1906.00544}}
  (\bibinfo{year}{2019}).
\newblock


\bibitem[Ha et~al\mbox{.}(2017)]%
        {david2017hypernetworks}
\bibfield{author}{\bibinfo{person}{David Ha}, \bibinfo{person}{Andrew Dai},
  {and} \bibinfo{person}{Quoc~V. Le}.} \bibinfo{year}{2017}\natexlab{}.
\newblock \showarticletitle{HyperNetworks}. In \bibinfo{booktitle}{\emph{Proc.
  ICLR}}.
\newblock


\bibitem[Han et~al\mbox{.}(2017)]%
        {han2017deepsketch2face}
\bibfield{author}{\bibinfo{person}{Xiaoguang Han}, \bibinfo{person}{Chang Gao},
  {and} \bibinfo{person}{Yizhou Yu}.} \bibinfo{year}{2017}\natexlab{}.
\newblock \showarticletitle{DeepSketch2Face: a deep learning based sketching
  system for 3D face and caricature modeling}.
\newblock \bibinfo{journal}{\emph{ACM Trans. Graph.}} \bibinfo{volume}{36},
  \bibinfo{number}{4} (\bibinfo{year}{2017}), \bibinfo{pages}{126}.
\newblock


\bibitem[Han et~al\mbox{.}(2018)]%
        {han2018caricatureshop}
\bibfield{author}{\bibinfo{person}{Xiaoguang Han}, \bibinfo{person}{Kangcheng
  Hou}, \bibinfo{person}{Dong Du}, \bibinfo{person}{Yuda Qiu},
  \bibinfo{person}{Shuguang Cui}, \bibinfo{person}{Kun Zhou}, {and}
  \bibinfo{person}{Yizhou Yu}.} \bibinfo{year}{2018}\natexlab{}.
\newblock \showarticletitle{Caricatureshop: Personalized and photorealistic
  caricature sketching}.
\newblock \bibinfo{journal}{\emph{IEEE Trans. Vis. Comput. Graph.}}
  \bibinfo{volume}{26}, \bibinfo{number}{7} (\bibinfo{year}{2018}),
  \bibinfo{pages}{2349--2361}.
\newblock


\bibitem[Jang et~al\mbox{.}(2021)]%
        {jang2021stylecarigan}
\bibfield{author}{\bibinfo{person}{Wonjong Jang}, \bibinfo{person}{Gwangjin
  Ju}, \bibinfo{person}{Yucheol Jung}, \bibinfo{person}{Jiaolong Yang},
  \bibinfo{person}{Xin Tong}, {and} \bibinfo{person}{Seungyong Lee}.}
  \bibinfo{year}{2021}\natexlab{}.
\newblock \showarticletitle{StyleCariGAN: caricature generation via StyleGAN
  feature map modulation}.
\newblock \bibinfo{journal}{\emph{ACM Trans. Graph.}} \bibinfo{volume}{40},
  \bibinfo{number}{4} (\bibinfo{year}{2021}), \bibinfo{pages}{1--16}.
\newblock


\bibitem[Ji et~al\mbox{.}(2020)]%
        {ji2020unsupervised}
\bibfield{author}{\bibinfo{person}{Wen Ji}, \bibinfo{person}{Kelei He},
  \bibinfo{person}{Jing Huo}, \bibinfo{person}{Zheng Gu}, {and}
  \bibinfo{person}{Yang Gao}.} \bibinfo{year}{2020}\natexlab{}.
\newblock \showarticletitle{Unsupervised domain attention adaptation network
  for caricature attribute recognition}. In \bibinfo{booktitle}{\emph{Proc.
  ECCV}}.
\newblock


\bibitem[Jiang et~al\mbox{.}(2019)]%
        {jiang2019disentangled}
\bibfield{author}{\bibinfo{person}{Zi-Hang Jiang}, \bibinfo{person}{Qianyi Wu},
  \bibinfo{person}{Keyu Chen}, {and} \bibinfo{person}{Juyong Zhang}.}
  \bibinfo{year}{2019}\natexlab{}.
\newblock \showarticletitle{Disentangled representation learning for 3D face
  shape}. In \bibinfo{booktitle}{\emph{Proc. CVPR}}.
\newblock


\bibitem[Karras et~al\mbox{.}(2019)]%
        {karras2019style}
\bibfield{author}{\bibinfo{person}{Tero Karras}, \bibinfo{person}{Samuli
  Laine}, {and} \bibinfo{person}{Timo Aila}.} \bibinfo{year}{2019}\natexlab{}.
\newblock \showarticletitle{A style-based generator architecture for generative
  adversarial networks}. In \bibinfo{booktitle}{\emph{Proc.\ CVPR}}.
\newblock


\bibitem[Liu et~al\mbox{.}(2022)]%
        {liu2022learning}
\bibfield{author}{\bibinfo{person}{Hsueh-Ti~Derek Liu},
  \bibinfo{person}{Francis Williams}, \bibinfo{person}{Alec Jacobson},
  \bibinfo{person}{Sanja Fidler}, {and} \bibinfo{person}{Or Litany}.}
  \bibinfo{year}{2022}\natexlab{}.
\newblock \showarticletitle{Learning Smooth Neural Functions via Lipschitz
  Regularization}.
\newblock \bibinfo{journal}{\emph{arXiv preprint arXiv:2202.08345}}
  (\bibinfo{year}{2022}).
\newblock


\bibitem[Liu et~al\mbox{.}(2020)]%
        {liu2020neural}
\bibfield{author}{\bibinfo{person}{Lingjie Liu}, \bibinfo{person}{Jiatao Gu},
  \bibinfo{person}{Kyaw Zaw~Lin}, \bibinfo{person}{Tat-Seng Chua}, {and}
  \bibinfo{person}{Christian Theobalt}.} \bibinfo{year}{2020}\natexlab{}.
\newblock \showarticletitle{Neural sparse voxel fields}.
\newblock \bibinfo{journal}{\emph{Advances in Neural Information Processing
  Systems}}  \bibinfo{volume}{33} (\bibinfo{year}{2020}),
  \bibinfo{pages}{15651--15663}.
\newblock


\bibitem[Lombardi et~al\mbox{.}(2018)]%
        {lombardi2018deep}
\bibfield{author}{\bibinfo{person}{Stephen Lombardi}, \bibinfo{person}{Jason
  Saragih}, \bibinfo{person}{Tomas Simon}, {and} \bibinfo{person}{Yaser
  Sheikh}.} \bibinfo{year}{2018}\natexlab{}.
\newblock \showarticletitle{Deep appearance models for face rendering}.
\newblock \bibinfo{journal}{\emph{ACM Transactions on Graphics (ToG)}}
  \bibinfo{volume}{37}, \bibinfo{number}{4} (\bibinfo{year}{2018}),
  \bibinfo{pages}{1--13}.
\newblock


\bibitem[Lombardi et~al\mbox{.}(2019)]%
        {lombardi2019neural}
\bibfield{author}{\bibinfo{person}{Stephen Lombardi}, \bibinfo{person}{Tomas
  Simon}, \bibinfo{person}{Jason Saragih}, \bibinfo{person}{Gabriel Schwartz},
  \bibinfo{person}{Andreas Lehrmann}, {and} \bibinfo{person}{Yaser Sheikh}.}
  \bibinfo{year}{2019}\natexlab{}.
\newblock \showarticletitle{Neural volumes: Learning dynamic renderable volumes
  from images}.
\newblock \bibinfo{journal}{\emph{arXiv preprint arXiv:1906.07751}}
  (\bibinfo{year}{2019}).
\newblock


\bibitem[Loop(1987)]%
        {loop1987smooth}
\bibfield{author}{\bibinfo{person}{Charles Loop}.}
  \bibinfo{year}{1987}\natexlab{}.
\newblock \emph{\bibinfo{title}{Smooth Subdivision Surfaces Based on
  Triangles}}.
\newblock \bibinfo{thesistype}{Ph.\,D. Dissertation}.
\newblock


\bibitem[L{\"u}thi et~al\mbox{.}(2017)]%
        {luthi2017gaussian}
\bibfield{author}{\bibinfo{person}{Marcel L{\"u}thi}, \bibinfo{person}{Thomas
  Gerig}, \bibinfo{person}{Christoph Jud}, {and} \bibinfo{person}{Thomas
  Vetter}.} \bibinfo{year}{2017}\natexlab{}.
\newblock \showarticletitle{Gaussian process morphable models}.
\newblock \bibinfo{journal}{\emph{IEEE Trans. PAMI}} \bibinfo{volume}{40},
  \bibinfo{number}{8} (\bibinfo{year}{2017}), \bibinfo{pages}{1860--1873}.
\newblock


\bibitem[Mildenhall et~al\mbox{.}(2020)]%
        {mildenhall2020nerf}
\bibfield{author}{\bibinfo{person}{Ben Mildenhall}, \bibinfo{person}{Pratul~P
  Srinivasan}, \bibinfo{person}{Matthew Tancik}, \bibinfo{person}{Jonathan~T
  Barron}, \bibinfo{person}{Ravi Ramamoorthi}, {and} \bibinfo{person}{Ren Ng}.}
  \bibinfo{year}{2020}\natexlab{}.
\newblock \showarticletitle{Nerf: Representing scenes as neural radiance fields
  for view synthesis}. In \bibinfo{booktitle}{\emph{Proc. ECCV}}.
\newblock


\bibitem[Park et~al\mbox{.}(2019)]%
        {park2019deepsdf}
\bibfield{author}{\bibinfo{person}{Jeong~Joon Park}, \bibinfo{person}{Peter
  Florence}, \bibinfo{person}{Julian Straub}, \bibinfo{person}{Richard
  Newcombe}, {and} \bibinfo{person}{Steven Lovegrove}.}
  \bibinfo{year}{2019}\natexlab{}.
\newblock \showarticletitle{Deepsdf: Learning continuous signed distance
  functions for shape representation}. In \bibinfo{booktitle}{\emph{Proc.
  CVPR}}.
\newblock


\bibitem[Paszke et~al\mbox{.}(2019)]%
        {NEURIPS2019_9015}
\bibfield{author}{\bibinfo{person}{Adam Paszke}, \bibinfo{person}{Sam Gross},
  \bibinfo{person}{Francisco Massa}, \bibinfo{person}{Adam Lerer},
  \bibinfo{person}{James Bradbury}, \bibinfo{person}{Gregory Chanan},
  \bibinfo{person}{Trevor Killeen}, \bibinfo{person}{Zeming Lin},
  \bibinfo{person}{Natalia Gimelshein}, \bibinfo{person}{Luca Antiga},
  \bibinfo{person}{Alban Desmaison}, \bibinfo{person}{Andreas Kopf},
  \bibinfo{person}{Edward Yang}, \bibinfo{person}{Zachary DeVito},
  \bibinfo{person}{Martin Raison}, \bibinfo{person}{Alykhan Tejani},
  \bibinfo{person}{Sasank Chilamkurthy}, \bibinfo{person}{Benoit Steiner},
  \bibinfo{person}{Lu Fang}, \bibinfo{person}{Junjie Bai}, {and}
  \bibinfo{person}{Soumith Chintala}.} \bibinfo{year}{2019}\natexlab{}.
\newblock \showarticletitle{PyTorch: An Imperative Style, High-Performance Deep
  Learning Library}.
\newblock In \bibinfo{booktitle}{\emph{Advances in Neural Information
  Processing Systems 32}}, \bibfield{editor}{\bibinfo{person}{H.~Wallach},
  \bibinfo{person}{H.~Larochelle}, \bibinfo{person}{A.~Beygelzimer},
  \bibinfo{person}{F.~d\textquotesingle Alch\'{e}-Buc},
  \bibinfo{person}{E.~Fox}, {and} \bibinfo{person}{R.~Garnett}} (Eds.).
  \bibinfo{publisher}{Curran Associates, Inc.}, \bibinfo{pages}{8024--8035}.
\newblock
\urldef\tempurl%
\url{http://papers.neurips.cc/paper/9015-pytorch-an-imperative-style-high-performance-deep-learning-library.pdf}
\showURL{%
\tempurl}


\bibitem[Ploumpis et~al\mbox{.}(2020)]%
        {ploumpis2020towards}
\bibfield{author}{\bibinfo{person}{Stylianos Ploumpis},
  \bibinfo{person}{Evangelos Ververas}, \bibinfo{person}{Eimear O'Sullivan},
  \bibinfo{person}{Stylianos Moschoglou}, \bibinfo{person}{Haoyang Wang},
  \bibinfo{person}{Nick Pears}, \bibinfo{person}{William~AP Smith},
  \bibinfo{person}{Baris Gecer}, {and} \bibinfo{person}{Stefanos Zafeiriou}.}
  \bibinfo{year}{2020}\natexlab{}.
\newblock \showarticletitle{Towards a complete 3D morphable model of the human
  head}.
\newblock \bibinfo{journal}{\emph{IEEE Trans. PAMI}} \bibinfo{volume}{43},
  \bibinfo{number}{11} (\bibinfo{year}{2020}), \bibinfo{pages}{4142--4160}.
\newblock


\bibitem[Qiu et~al\mbox{.}(2021)]%
        {qiu20213dcaricshop}
\bibfield{author}{\bibinfo{person}{Yuda Qiu}, \bibinfo{person}{Xiaojie Xu},
  \bibinfo{person}{Lingteng Qiu}, \bibinfo{person}{Yan Pan},
  \bibinfo{person}{Yushuang Wu}, \bibinfo{person}{Weikai Chen}, {and}
  \bibinfo{person}{Xiaoguang Han}.} \bibinfo{year}{2021}\natexlab{}.
\newblock \showarticletitle{3DCaricShop: A Dataset and A Baseline Method for
  Single-view 3D Caricature Face Reconstruction}. In
  \bibinfo{booktitle}{\emph{Proc. CVPR}}.
\newblock


\bibitem[Ranjan et~al\mbox{.}(2018)]%
        {ranjan2018generating}
\bibfield{author}{\bibinfo{person}{Anurag Ranjan}, \bibinfo{person}{Timo
  Bolkart}, \bibinfo{person}{Soubhik Sanyal}, {and} \bibinfo{person}{Michael~J
  Black}.} \bibinfo{year}{2018}\natexlab{}.
\newblock \showarticletitle{Generating 3D faces using convolutional mesh
  autoencoders}. In \bibinfo{booktitle}{\emph{Proc. ECCV}}.
\newblock


\bibitem[Saito et~al\mbox{.}(2019)]%
        {saito2019pifu}
\bibfield{author}{\bibinfo{person}{Shunsuke Saito}, \bibinfo{person}{Zeng
  Huang}, \bibinfo{person}{Ryota Natsume}, \bibinfo{person}{Shigeo Morishima},
  \bibinfo{person}{Angjoo Kanazawa}, {and} \bibinfo{person}{Hao Li}.}
  \bibinfo{year}{2019}\natexlab{}.
\newblock \showarticletitle{Pifu: Pixel-aligned implicit function for
  high-resolution clothed human digitization}. In
  \bibinfo{booktitle}{\emph{Proc. ICCV}}.
\newblock


\bibitem[Schwarz et~al\mbox{.}(2020)]%
        {schwarz2020graf}
\bibfield{author}{\bibinfo{person}{Katja Schwarz}, \bibinfo{person}{Yiyi Liao},
  \bibinfo{person}{Michael Niemeyer}, {and} \bibinfo{person}{Andreas Geiger}.}
  \bibinfo{year}{2020}\natexlab{}.
\newblock \showarticletitle{Graf: Generative radiance fields for 3d-aware image
  synthesis}.
\newblock \bibinfo{journal}{\emph{Advances in Neural Information Processing
  Systems}}  \bibinfo{volume}{33} (\bibinfo{year}{2020}),
  \bibinfo{pages}{20154--20166}.
\newblock


\bibitem[Sela et~al\mbox{.}(2015)]%
        {sela2015computational}
\bibfield{author}{\bibinfo{person}{Matan Sela}, \bibinfo{person}{Yonathan
  Aflalo}, {and} \bibinfo{person}{Ron Kimmel}.}
  \bibinfo{year}{2015}\natexlab{}.
\newblock \showarticletitle{Computational caricaturization of surfaces}.
\newblock \bibinfo{journal}{\emph{Computer Vision and Image Understanding}}
  \bibinfo{volume}{141} (\bibinfo{year}{2015}), \bibinfo{pages}{1--17}.
\newblock


\bibitem[Shen et~al\mbox{.}(2020)]%
        {shen2020interfacegan}
\bibfield{author}{\bibinfo{person}{Yujun Shen}, \bibinfo{person}{Ceyuan Yang},
  \bibinfo{person}{Xiaoou Tang}, {and} \bibinfo{person}{Bolei Zhou}.}
  \bibinfo{year}{2020}\natexlab{}.
\newblock \showarticletitle{Interfacegan: Interpreting the disentangled face
  representation learned by gans}.
\newblock \bibinfo{journal}{\emph{IEEE Trans. Pattern Anal. Mach. Intell.}}
  (\bibinfo{year}{2020}).
\newblock


\bibitem[Sitzmann et~al\mbox{.}(2020)]%
        {sitzmann2020implicit}
\bibfield{author}{\bibinfo{person}{Vincent Sitzmann},
  \bibinfo{person}{Julien~NP Martel}, \bibinfo{person}{Alexander~W Bergman},
  \bibinfo{person}{David~B Lindell}, {and} \bibinfo{person}{Gordon Wetzstein}.}
  \bibinfo{year}{2020}\natexlab{}.
\newblock \showarticletitle{Implicit neural representations with periodic
  activation functions}.
\newblock \bibinfo{journal}{\emph{Advances in Neural Information Processing
  Systems}}  \bibinfo{volume}{33} (\bibinfo{year}{2020}),
  \bibinfo{pages}{7462--7473}.
\newblock


\bibitem[Sitzmann et~al\mbox{.}(2019)]%
        {sitzmann2019scene}
\bibfield{author}{\bibinfo{person}{Vincent Sitzmann}, \bibinfo{person}{Michael
  Zollh{\"o}fer}, {and} \bibinfo{person}{Gordon Wetzstein}.}
  \bibinfo{year}{2019}\natexlab{}.
\newblock \showarticletitle{Scene representation networks: Continuous
  3d-structure-aware neural scene representations}.
\newblock \bibinfo{journal}{\emph{Advances in Neural Information Processing
  Systems}}  \bibinfo{volume}{32} (\bibinfo{year}{2019}).
\newblock


\bibitem[Sorkine and Alexa(2007)]%
        {sorkine2007as}
\bibfield{author}{\bibinfo{person}{Olga Sorkine} {and} \bibinfo{person}{Marc
  Alexa}.} \bibinfo{year}{2007}\natexlab{}.
\newblock \showarticletitle{As-rigid-as-possible surface modeling}. In
  \bibinfo{booktitle}{\emph{Proc. SGP}}.
\newblock


\bibitem[Thies et~al\mbox{.}(2016)]%
        {thies2016face2face}
\bibfield{author}{\bibinfo{person}{Justus Thies}, \bibinfo{person}{Michael
  Zollhofer}, \bibinfo{person}{Marc Stamminger}, \bibinfo{person}{Christian
  Theobalt}, {and} \bibinfo{person}{Matthias Nie{\ss}ner}.}
  \bibinfo{year}{2016}\natexlab{}.
\newblock \showarticletitle{Face2face: Real-time face capture and reenactment
  of rgb videos}. In \bibinfo{booktitle}{\emph{Proc. CVPR}}.
\newblock


\bibitem[Wang et~al\mbox{.}(2019)]%
        {wang20193dn}
\bibfield{author}{\bibinfo{person}{Weiyue Wang}, \bibinfo{person}{Duygu
  Ceylan}, \bibinfo{person}{Radomir Mech}, {and} \bibinfo{person}{Ulrich
  Neumann}.} \bibinfo{year}{2019}\natexlab{}.
\newblock \showarticletitle{3dn: 3d deformation network}. In
  \bibinfo{booktitle}{\emph{Proc. CVPR}}.
\newblock


\bibitem[Wu et~al\mbox{.}(2018)]%
        {wu2018alive}
\bibfield{author}{\bibinfo{person}{Qianyi Wu}, \bibinfo{person}{Juyong Zhang},
  \bibinfo{person}{Yu-Kun Lai}, \bibinfo{person}{Jianmin Zheng}, {and}
  \bibinfo{person}{Jianfei Cai}.} \bibinfo{year}{2018}\natexlab{}.
\newblock \showarticletitle{Alive Caricature from 2D to 3D}. In
  \bibinfo{booktitle}{\emph{Proc. CVPR}}.
\newblock


\end{thebibliography}



\begin{thebibliography}{18}


\ifx \showCODEN    \undefined \def \showCODEN     #1{\unskip}     \fi
\ifx \showDOI      \undefined \def \showDOI       #1{#1}\fi
\ifx \showISBNx    \undefined \def \showISBNx     #1{\unskip}     \fi
\ifx \showISBNxiii \undefined \def \showISBNxiii  #1{\unskip}     \fi
\ifx \showISSN     \undefined \def \showISSN      #1{\unskip}     \fi
\ifx \showLCCN     \undefined \def \showLCCN      #1{\unskip}     \fi
\ifx \shownote     \undefined \def \shownote      #1{#1}          \fi
\ifx \showarticletitle \undefined \def \showarticletitle #1{#1}   \fi
\ifx \showURL      \undefined \def \showURL       {\relax}        \fi
\providecommand\bibfield[2]{#2}
\providecommand\bibinfo[2]{#2}
\providecommand\natexlab[1]{#1}
\providecommand\showeprint[2][]{arXiv:#2}

\bibitem[Bogo et~al\mbox{.}(2017)]%
        {bogo2017dfaust}
\bibfield{author}{\bibinfo{person}{Federica Bogo}, \bibinfo{person}{Javier
  Romero}, \bibinfo{person}{Gerard Pons-Moll}, {and}
  \bibinfo{person}{Michael~J. Black}.} \bibinfo{year}{2017}\natexlab{}.
\newblock \showarticletitle{Dynamic {FAUST}: {R}egistering Human Bodies in
  Motion}. In \bibinfo{booktitle}{\emph{Proc. CVPR}}.
\newblock


\bibitem[Cao et~al\mbox{.}(2013)]%
        {cao2013facewarehouse}
\bibfield{author}{\bibinfo{person}{Chen Cao}, \bibinfo{person}{Yanlin Weng},
  \bibinfo{person}{Shun Zhou}, \bibinfo{person}{Yiying Tong}, {and}
  \bibinfo{person}{Kun Zhou}.} \bibinfo{year}{2013}\natexlab{}.
\newblock \showarticletitle{Facewarehouse: A 3d facial expression database for
  visual computing}.
\newblock \bibinfo{journal}{\emph{IEEE Trans. Vis. Comput. Graph.}}
  \bibinfo{volume}{20}, \bibinfo{number}{3} (\bibinfo{year}{2013}),
  \bibinfo{pages}{413--425}.
\newblock


\bibitem[Cortes and Vapnik(1995)]%
        {cortes1995support}
\bibfield{author}{\bibinfo{person}{Corinna Cortes} {and}
  \bibinfo{person}{Vladimir Vapnik}.} \bibinfo{year}{1995}\natexlab{}.
\newblock \showarticletitle{Support-vector networks}.
\newblock \bibinfo{journal}{\emph{Machine learning}} \bibinfo{volume}{20},
  \bibinfo{number}{3} (\bibinfo{year}{1995}), \bibinfo{pages}{273--297}.
\newblock


\bibitem[Guo et~al\mbox{.}(2019)]%
        {guo20193d}
\bibfield{author}{\bibinfo{person}{Yudong Guo}, \bibinfo{person}{Luo Jiang},
  \bibinfo{person}{Lin Cai}, {and} \bibinfo{person}{Juyong Zhang}.}
  \bibinfo{year}{2019}\natexlab{}.
\newblock \showarticletitle{3D Magic Mirror: Automatic Video to 3D Caricature
  Translation}.
\newblock \bibinfo{journal}{\emph{arXiv preprint arXiv:1906.00544}}
  (\bibinfo{year}{2019}).
\newblock


\bibitem[Jacobson et~al\mbox{.}(2018)]%
        {libigl}
\bibfield{author}{\bibinfo{person}{Alec Jacobson}, \bibinfo{person}{Daniele
  Panozzo}, {et~al\mbox{.}}} \bibinfo{year}{2018}\natexlab{}.
\newblock \bibinfo{title}{{libigl}: A simple {C++} geometry processing
  library}.
\newblock
\newblock
\newblock
\shownote{http://libigl.github.io/libigl/}.


\bibitem[Ji et~al\mbox{.}(2020)]%
        {ji2020unsupervised}
\bibfield{author}{\bibinfo{person}{Wen Ji}, \bibinfo{person}{Kelei He},
  \bibinfo{person}{Jing Huo}, \bibinfo{person}{Zheng Gu}, {and}
  \bibinfo{person}{Yang Gao}.} \bibinfo{year}{2020}\natexlab{}.
\newblock \showarticletitle{Unsupervised domain attention adaptation network
  for caricature attribute recognition}. In \bibinfo{booktitle}{\emph{Proc.
  ECCV}}.
\newblock


\bibitem[Jiang et~al\mbox{.}(2018)]%
        {jiang20183d}
\bibfield{author}{\bibinfo{person}{Luo Jiang}, \bibinfo{person}{Juyong Zhang},
  \bibinfo{person}{Bailin Deng}, \bibinfo{person}{Hao Li}, {and}
  \bibinfo{person}{Ligang Liu}.} \bibinfo{year}{2018}\natexlab{}.
\newblock \showarticletitle{3D face reconstruction with geometry details from a
  single image}.
\newblock \bibinfo{journal}{\emph{IEEE Trans. Image Process.}}
  \bibinfo{volume}{27}, \bibinfo{number}{10} (\bibinfo{year}{2018}),
  \bibinfo{pages}{4756--4770}.
\newblock


\bibitem[Jiang et~al\mbox{.}(2019)]%
        {jiang2019disentangled}
\bibfield{author}{\bibinfo{person}{Zi-Hang Jiang}, \bibinfo{person}{Qianyi Wu},
  \bibinfo{person}{Keyu Chen}, {and} \bibinfo{person}{Juyong Zhang}.}
  \bibinfo{year}{2019}\natexlab{}.
\newblock \showarticletitle{Disentangled representation learning for 3D face
  shape}. In \bibinfo{booktitle}{\emph{Proc. CVPR}}.
\newblock


\bibitem[Kemelmacher-Shlizerman and Seitz(2011)]%
        {kemelmacher2011face}
\bibfield{author}{\bibinfo{person}{Ira Kemelmacher-Shlizerman} {and}
  \bibinfo{person}{Steven~M Seitz}.} \bibinfo{year}{2011}\natexlab{}.
\newblock \showarticletitle{Face reconstruction in the wild}. In
  \bibinfo{booktitle}{\emph{2011 international conference on computer vision}}.
  IEEE, \bibinfo{pages}{1746--1753}.
\newblock


\bibitem[Kingma and Ba(2015)]%
        {kingma2015adam}
\bibfield{author}{\bibinfo{person}{Diederik~P Kingma} {and}
  \bibinfo{person}{Jimmy Ba}.} \bibinfo{year}{2015}\natexlab{}.
\newblock \showarticletitle{Adam: A Method for Stochastic Optimization}. In
  \bibinfo{booktitle}{\emph{Proc. ICLR}}.
\newblock


\bibitem[Pearson(1901)]%
        {pearson1901liii}
\bibfield{author}{\bibinfo{person}{Karl Pearson}.}
  \bibinfo{year}{1901}\natexlab{}.
\newblock \showarticletitle{LIII. On lines and planes of closest fit to systems
  of points in space}.
\newblock \bibinfo{journal}{\emph{The London, Edinburgh, and Dublin
  philosophical magazine and journal of science}} \bibinfo{volume}{2},
  \bibinfo{number}{11} (\bibinfo{year}{1901}), \bibinfo{pages}{559--572}.
\newblock


\bibitem[Qiu et~al\mbox{.}(2021)]%
        {qiu20213dcaricshop}
\bibfield{author}{\bibinfo{person}{Yuda Qiu}, \bibinfo{person}{Xiaojie Xu},
  \bibinfo{person}{Lingteng Qiu}, \bibinfo{person}{Yan Pan},
  \bibinfo{person}{Yushuang Wu}, \bibinfo{person}{Weikai Chen}, {and}
  \bibinfo{person}{Xiaoguang Han}.} \bibinfo{year}{2021}\natexlab{}.
\newblock \showarticletitle{3DCaricShop: A Dataset and A Baseline Method for
  Single-view 3D Caricature Face Reconstruction}. In
  \bibinfo{booktitle}{\emph{Proc. CVPR}}.
\newblock


\bibitem[Ranjan et~al\mbox{.}(2018)]%
        {ranjan2018generating}
\bibfield{author}{\bibinfo{person}{Anurag Ranjan}, \bibinfo{person}{Timo
  Bolkart}, \bibinfo{person}{Soubhik Sanyal}, {and} \bibinfo{person}{Michael~J
  Black}.} \bibinfo{year}{2018}\natexlab{}.
\newblock \showarticletitle{Generating 3D faces using convolutional mesh
  autoencoders}. In \bibinfo{booktitle}{\emph{Proc. ECCV}}.
\newblock


\bibitem[Russian3DScanner(2022)]%
        {wrap3}
\bibfield{author}{\bibinfo{person}{Russian3DScanner}.}
  \bibinfo{year}{2022}\natexlab{}.
\newblock \bibinfo{booktitle}{\emph{R3DS Wrap}}.
\newblock
\urldef\tempurl%
\url{https://www.russian3dscanner.com/}
\showURL{%
\tempurl}


\bibitem[Shen et~al\mbox{.}(2020)]%
        {shen2020interfacegan}
\bibfield{author}{\bibinfo{person}{Yujun Shen}, \bibinfo{person}{Ceyuan Yang},
  \bibinfo{person}{Xiaoou Tang}, {and} \bibinfo{person}{Bolei Zhou}.}
  \bibinfo{year}{2020}\natexlab{}.
\newblock \showarticletitle{Interfacegan: Interpreting the disentangled face
  representation learned by gans}.
\newblock \bibinfo{journal}{\emph{IEEE Trans. Pattern Anal. Mach. Intell.}}
  (\bibinfo{year}{2020}).
\newblock


\bibitem[Sitzmann et~al\mbox{.}(2020)]%
        {sitzmann2020implicit}
\bibfield{author}{\bibinfo{person}{Vincent Sitzmann},
  \bibinfo{person}{Julien~NP Martel}, \bibinfo{person}{Alexander~W Bergman},
  \bibinfo{person}{David~B Lindell}, {and} \bibinfo{person}{Gordon Wetzstein}.}
  \bibinfo{year}{2020}\natexlab{}.
\newblock \showarticletitle{Implicit neural representations with periodic
  activation functions}.
\newblock \bibinfo{journal}{\emph{Advances in Neural Information Processing
  Systems}}  \bibinfo{volume}{33} (\bibinfo{year}{2020}),
  \bibinfo{pages}{7462--7473}.
\newblock


\bibitem[Sorkine and Alexa(2007)]%
        {sorkine2007as}
\bibfield{author}{\bibinfo{person}{Olga Sorkine} {and} \bibinfo{person}{Marc
  Alexa}.} \bibinfo{year}{2007}\natexlab{}.
\newblock \showarticletitle{As-rigid-as-possible surface modeling}. In
  \bibinfo{booktitle}{\emph{Proc. SGP}}.
\newblock


\bibitem[Wu et~al\mbox{.}(2018)]%
        {wu2018alive}
\bibfield{author}{\bibinfo{person}{Qianyi Wu}, \bibinfo{person}{Juyong Zhang},
  \bibinfo{person}{Yu-Kun Lai}, \bibinfo{person}{Jianmin Zheng}, {and}
  \bibinfo{person}{Jianfei Cai}.} \bibinfo{year}{2018}\natexlab{}.
\newblock \showarticletitle{Alive Caricature from 2D to 3D}. In
  \bibinfo{booktitle}{\emph{Proc. CVPR}}.
\newblock


\end{thebibliography}

\end{document}


\author{Yucheol Jung}
\orcid{0000-0003-1593-4626}
\affiliation{%
  \institution{POSTECH}  
  \city{Pohang}
  \country{Republic of Korea}
  }
\email{ycjung@postech.ac.kr}

\author{Wonjong Jang}
\orcid{0000-0002-1442-9399}
\affiliation{%
  \institution{POSTECH} 
  \city{Pohang}
  \country{Republic of Korea}
  }
\email{wonjong@postech.ac.kr}

\author{Soongjin Kim}
\orcid{0000-0001-8142-7062}
\affiliation{%
  \institution{POSTECH} 
  \city{Pohang}
  \country{Republic of Korea}
  }
\email{kimsj0302@postech.ac.kr}

\author{Jiaolong Yang}
\orcid{0000-0002-7314-6567}
\affiliation{%
  \institution{Microsoft Research Asia}
  \city{Beijing}
  \country{China}
}
\email{jiaoyan@microsoft.com}

\author{Xin Tong}
\orcid{0000-0001-8788-2453}
\affiliation{%
  \institution{Microsoft Research Asia}
  \city{Beijing}
  \country{China}
}
\email{xtong@microsoft.com}

\author{Seungyong Lee}
\orcid{0000-0002-8159-4271}
\affiliation{%
  \institution{POSTECH}
  \city{Pohang}
  \country{Republic of Korea}
}
\email{leesy@postech.ac.kr}
\title{Supplementary material: \\ Deep Deformable 3D Caricatures with Learned Shape Control}
\maketitle
\title{Supplementary material: Deep Deformable 3D Caricatures with Learned Shape Control}

We provide details of the proposed method and more experimental results in this supplementary document. \Sec{dataset} provides quantitative analysis on the complexity of the 3DCaricShop dataset. \Sec{network} provides details on our training and the network structure. \Sec{detail-comparison} provides more details on Sec~4.1. in the main paper. \Sec{ablation-structure} provides visual results for the ablation study in the main paper.
\Sec{detail-landmark} provides details on the 3D caricature reconstruction from 2D landmarks. \Sec{interfacegan} provides details on our semantic face editing.
\Sec{arap} visually compares our point-handle-based editing to as-rigid-as-possible (ARAP) deformation \cite{sorkine2007as}. 
Lastly, \Sec{facewarehouse} explains data processing details in our automatic 3D caricature creation. For a large gallery of results using real-world caricatures and visual comparison with Alive Caricature \cite{wu2018alive} and 3DCaricShop \cite{qiu20213dcaricshop}, refer to our project page\footnote{  \href{https://ycjungsubhuman.github.io/DeepDeformable3DCaricatures}{https://ycjungsubhuman.github.io/DeepDeformable3DCaricatures}}. 

\section{Dataset analysis}
\label{sec:dataset}

Our key challenge in learning an editing space is the high variation and complexity of 3D caricatures. We observed the 3DCaricShop \cite{qiu20213dcaricshop} dataset contains much more complex information compared to other common 3D mesh datasets. We compare 3DCaricShop with a regular face dataset (FaceWarehouse \cite{cao2013facewarehouse}), a dynamic human body dataset (DFAUST \cite{bogo2017dfaust}), and a facial animation dataset (COMA \cite{ranjan2018generating}).

We evaluate the complexity of each dataset by observing the effectiveness of dimension reduction on the dataset. We apply principal component analysis (PCA) \cite{pearson1901liii} to each dataset and calculate the number of principal components required to explain 99\% of variations in the dataset. 3DCaricShop requires 178 principal components, which is a huge number compared to FaceWarehouse (52), DFAUST (22), and COMA (41). 3DCaricShop provides 2K examples, which are sparse samples compared to other datasets: FaceWarehouse (7K), DFAUST (41K), and COMA (20K). In addition, 3DCaricShop contains diverse high-frequency variations, while other datasets contain mostly low-frequency variations from shape, pose, or expression.

\section{Training details}
\label{sec:network}

Our model is optimized with the Adam optimizer \cite{kingma2015adam} using batch size 128, with learning rate $1.0 \times 10^{-4}$, $\beta_1=0.9$, and $\beta_2=0.999$.

\subsubsection{SIREN MLP} We construct the SIREN MLP with five fully connected layers: the first SIREN layer mapping a 3D position to a 128-dimension feature, three hidden SIREN layers with hidden feature dimension 128, and the last layer with feature dimension 3.

\subsubsection{hypernetwork} We have two ReLU MLPs for each fully-connected layer in the SIREN MLP: One for weights and the other for biases. Each ReLU MLP produces network parameters given a 128-dimension latent code. The ReLU MLPs have three fully connected layers each: the first two ReLU layers with output feature dimension 256, and the last fully connected layer without activation. For more details on hypernetworks, refer to \cite{sitzmann2020implicit}.

\subsubsection{Point sampling}

We sample two groups of points from a mesh; vertex position in the mesh and uniformly sampled points on the surface. Collating the two groups, we obtain 23,132 points for each example. The sampling is done statically before training. Each point $p_i$ on an example is mapped to the corresponding point $\hat{p}_i$ on the fixed template mesh by converting a vertex position on the example mesh into barycentric coordinates, then converting the barycentric coordinates to a template mesh position.

\section{Details on \textit{Comparison between different shape representation (Sec.\ 4.1)}}
\label{sec:detail-comparison}

The design of the compared model using the vertex position array is inspired by \cite{guo20193d} and \cite{jiang2019disentangled}. The model is a single MLP, and the training is done using the same auto-decoder architecture as ours. The MLP takes a 128-dimension latent code to produce an array of 3D vertex positions. The MLP is constructed using two fully connected layers with the following channel dimensions: $(128) \rightarrow (400) \rightarrow (3 \times 11551)$. The activation for the first layer is leaky ReLU with a negative slope of 0.1. The number of parameters is 13.9M, which is similar to that of ours with 14M parameters. The training is done using the same loss, optimizer, and batch size as our network.
We show the results of the compared model after approximately 17,500 epochs of training (8 hours). The result for our proposed network is from 1,500 epochs of training (2 hours). Both trainings were done on an Intel(R) Xeon(R) Silver 4114 server using two NVIDIA Titan Xp GPUs.
The reconstruction for the comparison using the test set is done by optimizing latent codes using the same loss used in training. In this optimization, we only use vertex positions on the mesh without random sampling on the surface. 

\section{Visual Results of Ablation Study}
\label{sec:ablation-structure}

\begin{figure*}[t]
    \centering
    \begin{tabular}{c@{\hspace{10pt}}c@{\hspace{5pt}}c@{}c@{}c@{}c}
    
    \vspace{-0.01cm}
    
    \includegraphics[align=c, width=0.15\textwidth, height=0.15\textwidth]{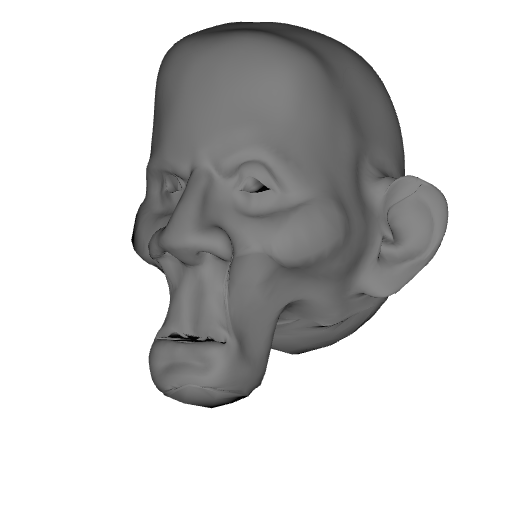} 
    &
    (a) &
    \includegraphics[align=c, width=0.15\textwidth, height=0.15\textwidth]{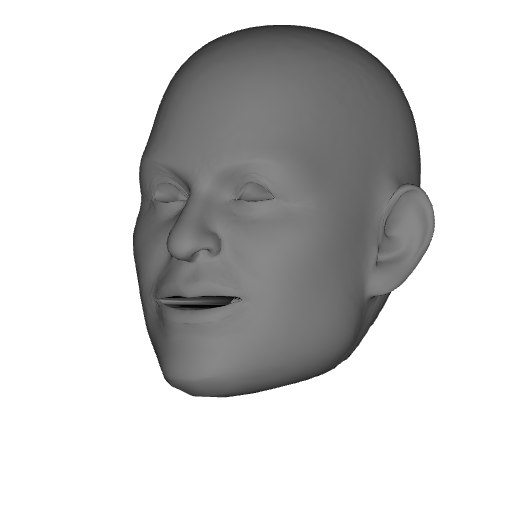} &
    \includegraphics[align=c, width=0.15\textwidth, height=0.15\textwidth]{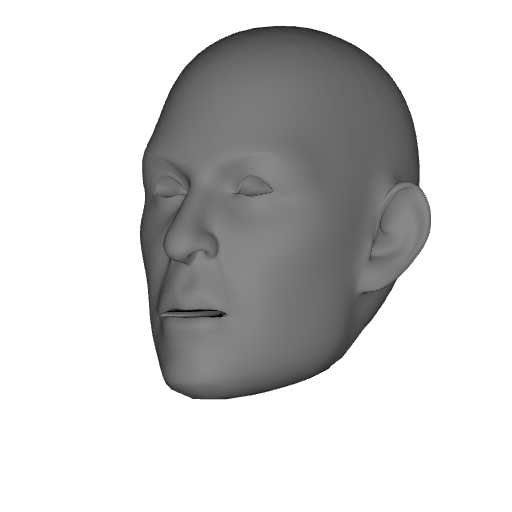} &
    \includegraphics[align=c, width=0.15\textwidth, height=0.15\textwidth]{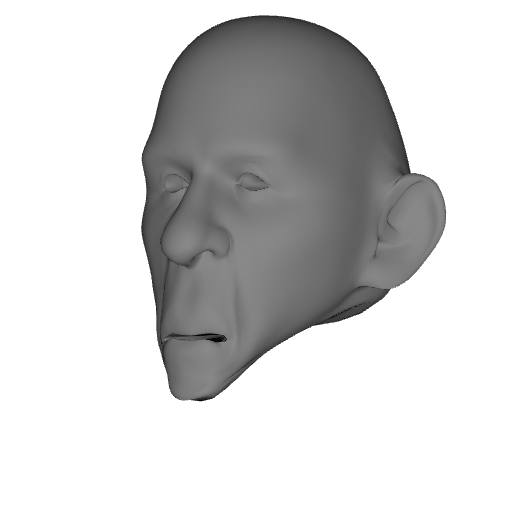} &
    \includegraphics[align=c, width=0.15\textwidth, height=0.15\textwidth]{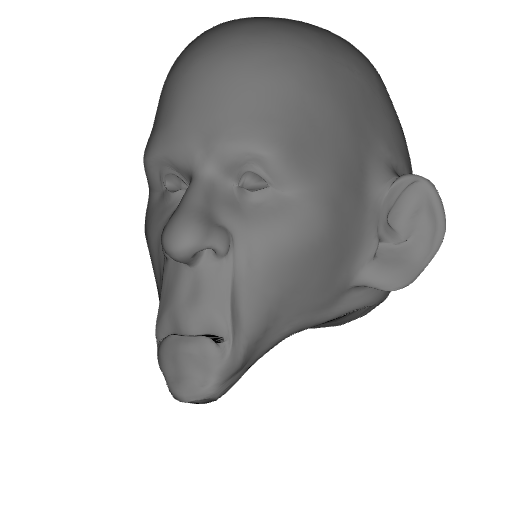} \\
    
    Ground truth & & & & &
    \vspace{-0.30cm} \\
     
    &
    (b) &
    \includegraphics[align=c, width=0.15\textwidth, height=0.15\textwidth]{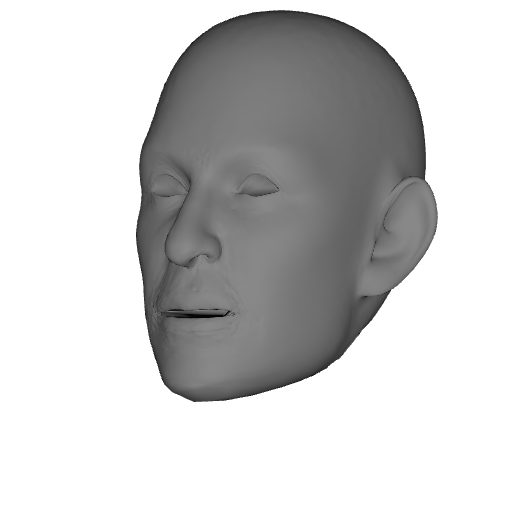} &
    \includegraphics[align=c, width=0.15\textwidth, height=0.15\textwidth]{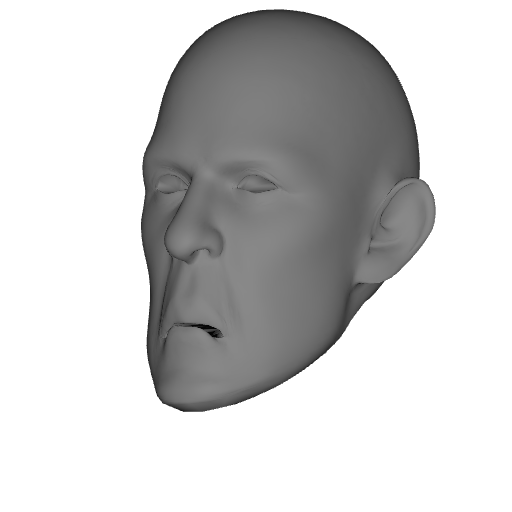} &
    \includegraphics[align=c, width=0.15\textwidth, height=0.15\textwidth]{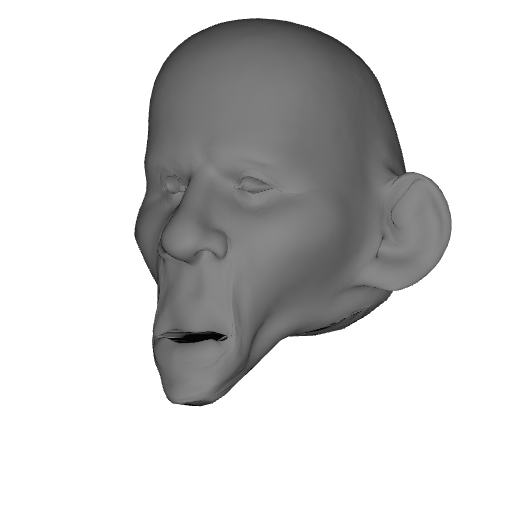} &
    \includegraphics[align=c, width=0.15\textwidth, height=0.15\textwidth]{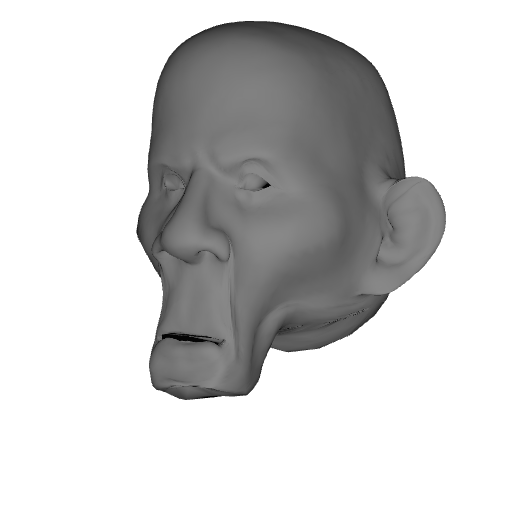} \\
    
     &
    (c) &
    \includegraphics[align=c, width=0.15\textwidth, height=0.15\textwidth]{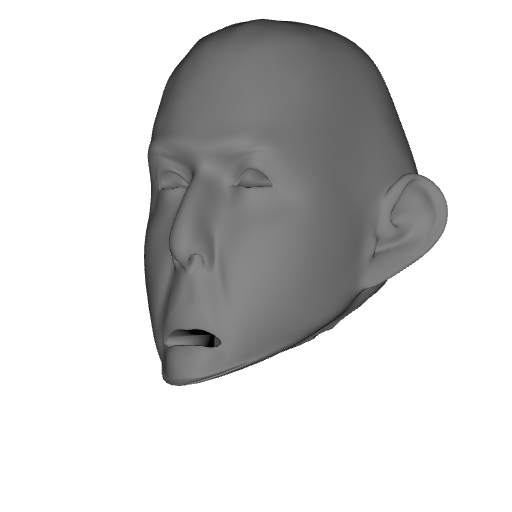} &
    \includegraphics[align=c, width=0.15\textwidth, height=0.15\textwidth]{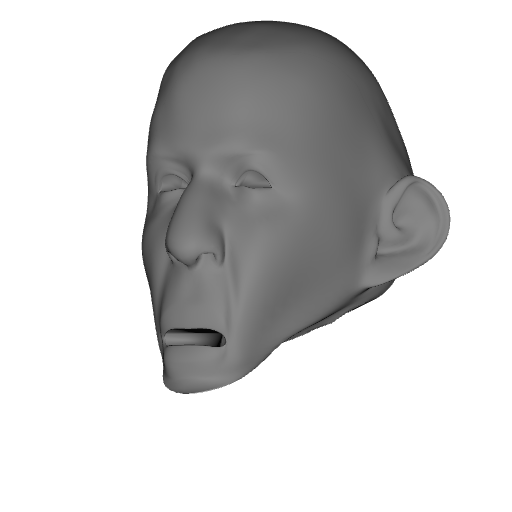} &
    \includegraphics[align=c, width=0.15\textwidth, height=0.15\textwidth]{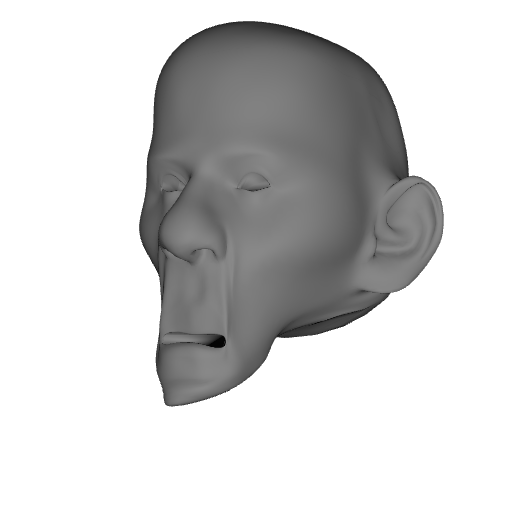} &
    \includegraphics[align=c, width=0.15\textwidth, height=0.15\textwidth]{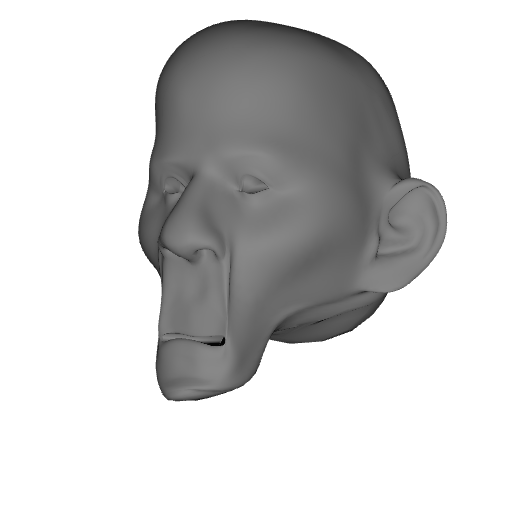} \\
    
    &
    (d) &
    \includegraphics[align=c, width=0.15\textwidth, height=0.15\textwidth]{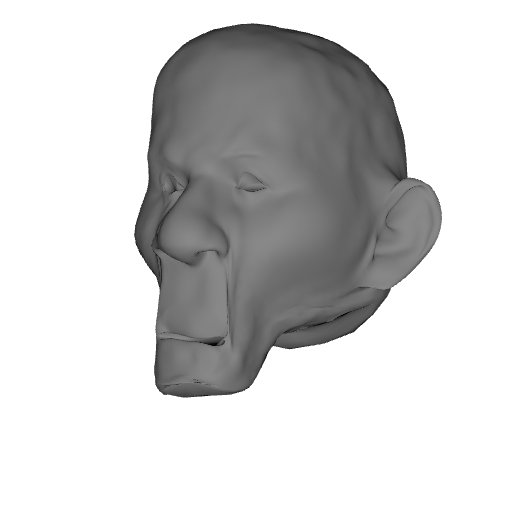} &
    \includegraphics[align=c, width=0.15\textwidth, height=0.15\textwidth]{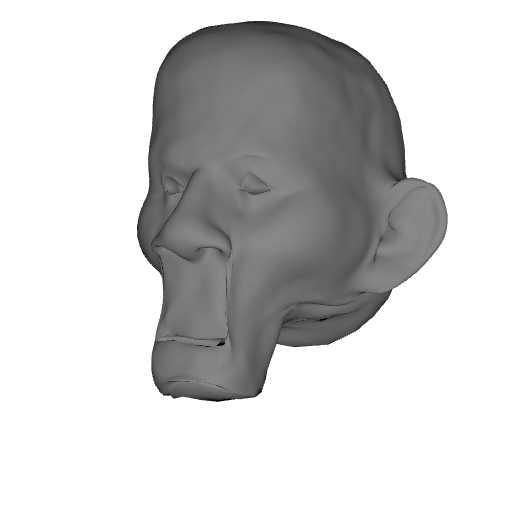} &
    \includegraphics[align=c, width=0.15\textwidth, height=0.15\textwidth]{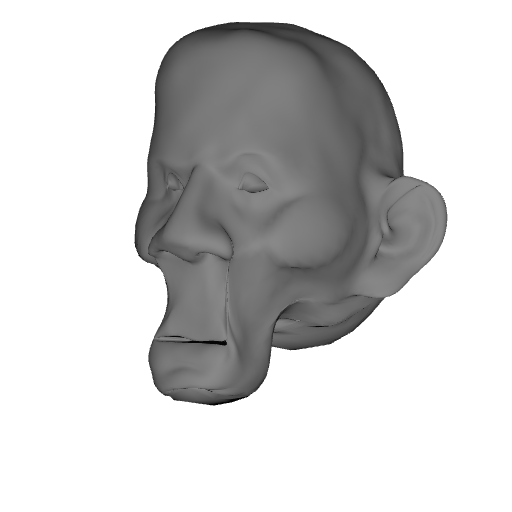} &
    \includegraphics[align=c, width=0.15\textwidth, height=0.15\textwidth]{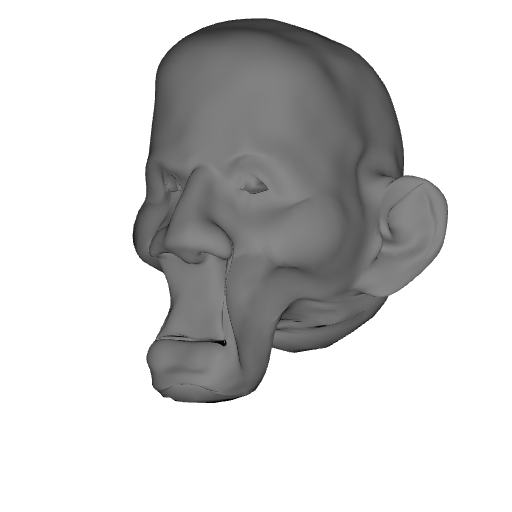} \\
    
     & & Epoch 500 & 1500 & 15000 & 29000 \\
    \end{tabular}
    \caption{Visual result of training set reconstruction for different model configurations. (a) Vertex position array MLP. (b) Vertex displacement array MLP. (c) SIREN MLP with conditioning. (d) SIREN MLP hypernetwork.}
    \label{fig:ablation-vis-train}
\end{figure*}

\begin{figure*}[t]
    \centering
    \begin{tabular}{c@{\hspace{10pt}}c@{\hspace{5pt}}c@{}c@{}c@{}c}
    
    \vspace{-0.01cm}
    \includegraphics[align=c, width=0.15\textwidth, height=0.15\textwidth]{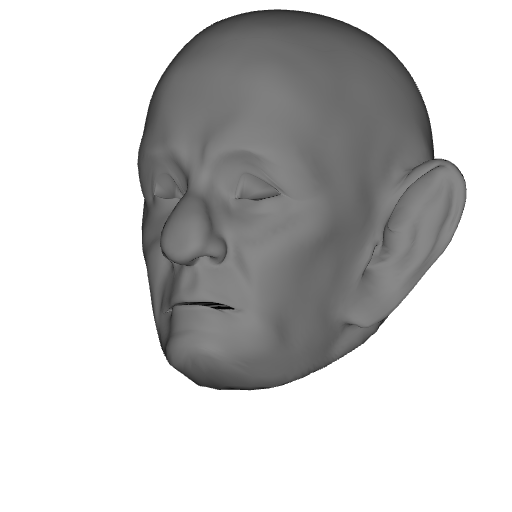}  &
    (a) &
    \includegraphics[align=c, width=0.15\textwidth, height=0.15\textwidth]{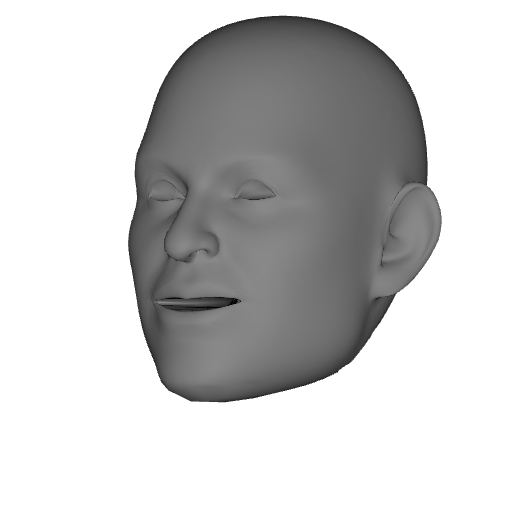} &
    \includegraphics[align=c, width=0.15\textwidth, height=0.15\textwidth]{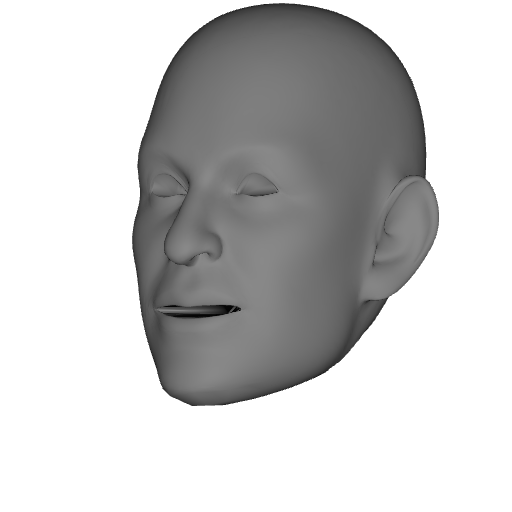} &
    \includegraphics[align=c, width=0.15\textwidth, height=0.15\textwidth]{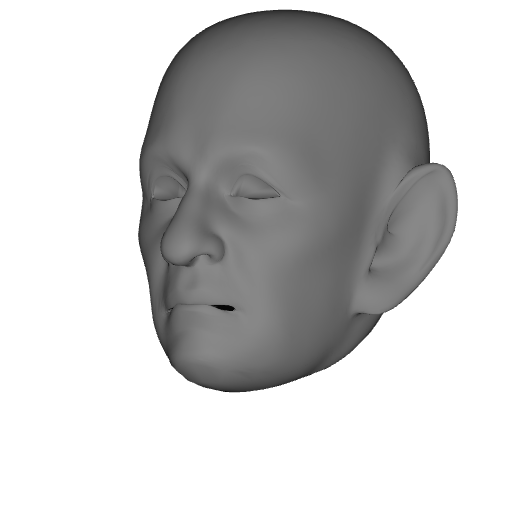} &
    \includegraphics[align=c, width=0.15\textwidth, height=0.15\textwidth]{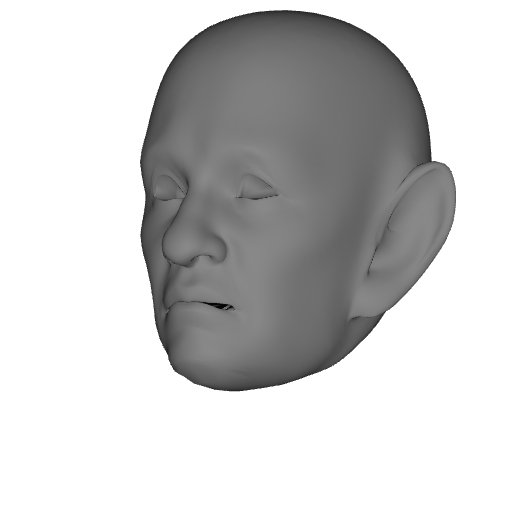} \\
    
    Ground truth & & & & &
    \vspace{-0.30cm} \\
    
     &
    (b) &
    \includegraphics[align=c, width=0.15\textwidth, height=0.15\textwidth]{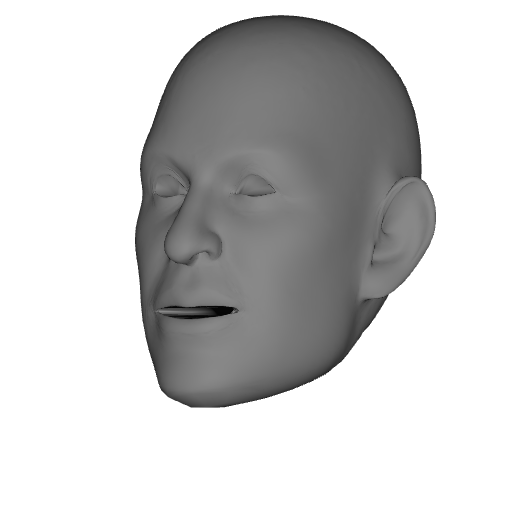} &
    \includegraphics[align=c, width=0.15\textwidth, height=0.15\textwidth]{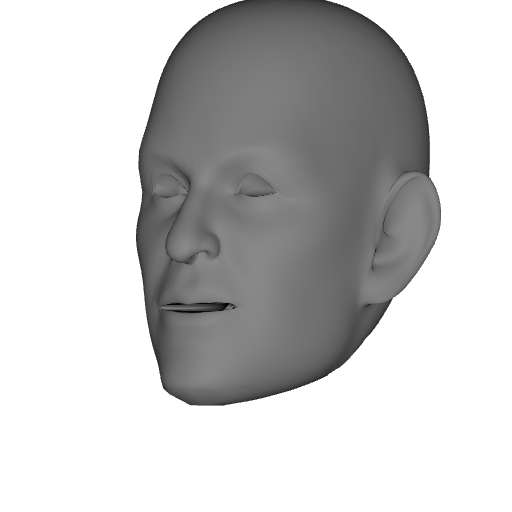} &
    \includegraphics[align=c, width=0.15\textwidth, height=0.15\textwidth]{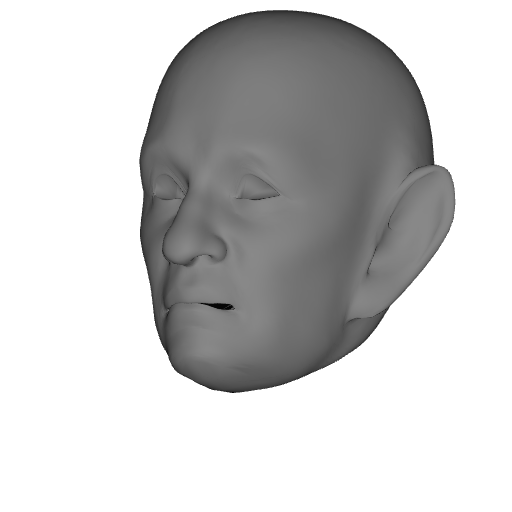} &
    \includegraphics[align=c, width=0.15\textwidth, height=0.15\textwidth]{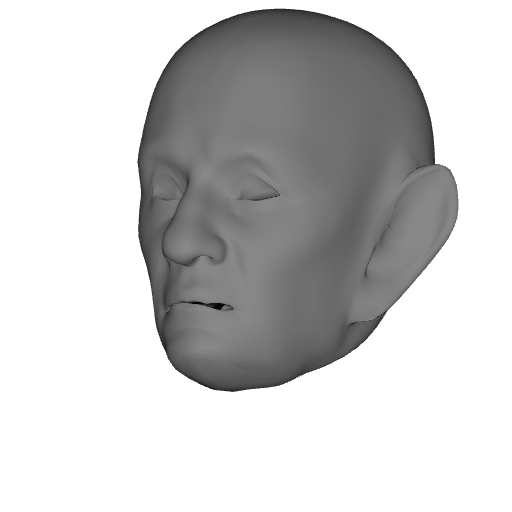} \\
    
     &
    (c) &
    \includegraphics[align=c, width=0.15\textwidth, height=0.15\textwidth]{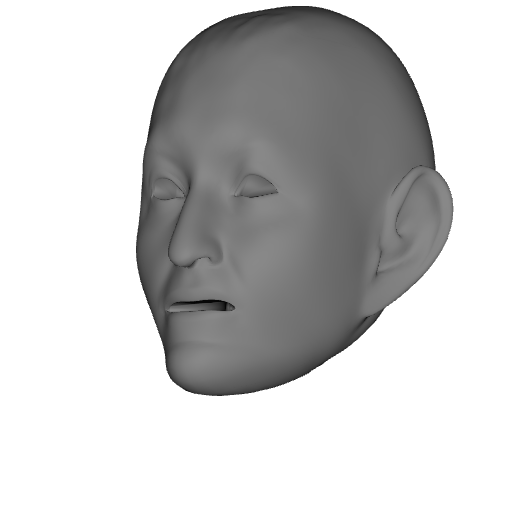} &
    \includegraphics[align=c, width=0.15\textwidth, height=0.15\textwidth]{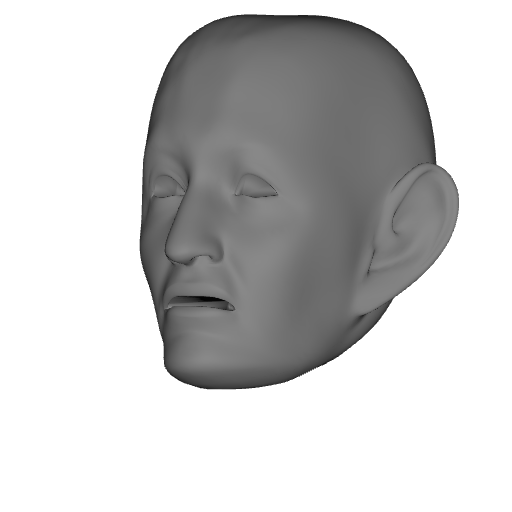} &
    \includegraphics[align=c, width=0.15\textwidth, height=0.15\textwidth]{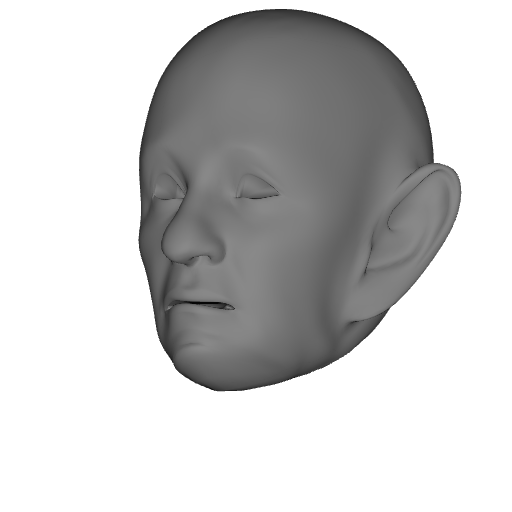} &
    \includegraphics[align=c, width=0.15\textwidth, height=0.15\textwidth]{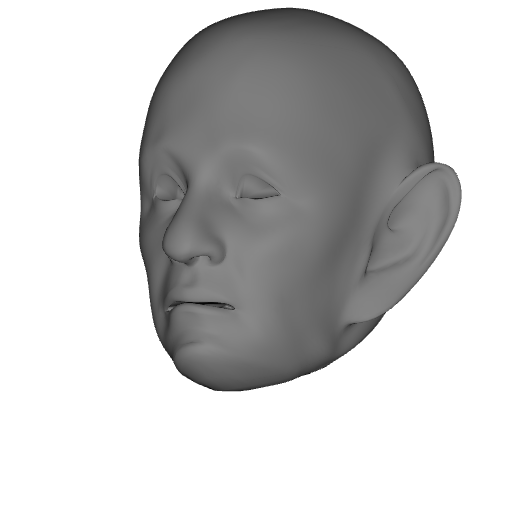} \\
    
    &
    (d) &
    \includegraphics[align=c, width=0.15\textwidth, height=0.15\textwidth]{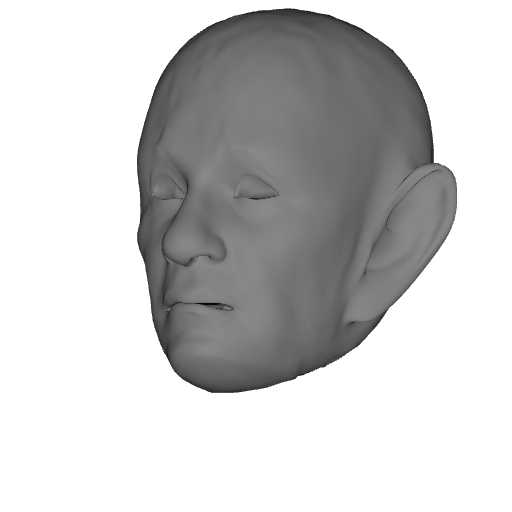} &
    \includegraphics[align=c, width=0.15\textwidth, height=0.15\textwidth]{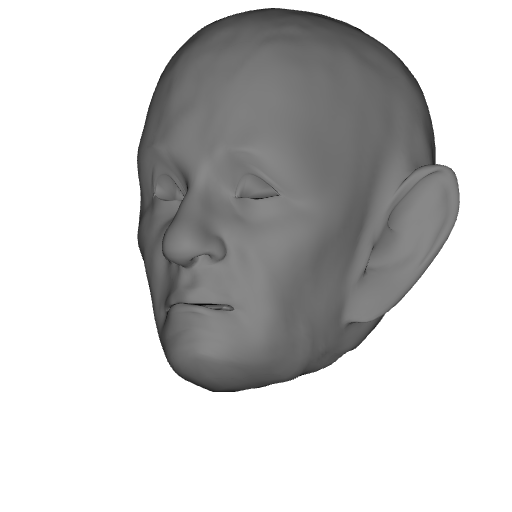} &
    \includegraphics[align=c, width=0.15\textwidth, height=0.15\textwidth]{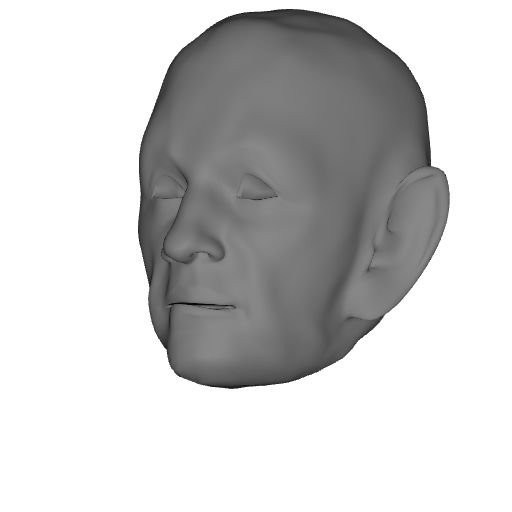} &
    \includegraphics[align=c, width=0.15\textwidth, height=0.15\textwidth]{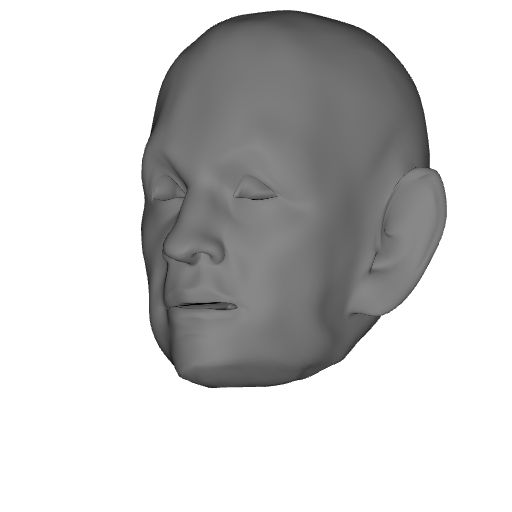} \\
    
     & & Epoch 500 & 1500 & 15000 & 29000 \\
    \end{tabular}
    \caption{Visual result of test set reconstruction for different model configurations. (a) Vertex position array MLP. (b) Vertex displacement array MLP. (c) SIREN MLP with conditioning. (d) SIREN MLP hypernetwork.}
    \label{fig:ablation-vis-test}
\end{figure*}

In addition to error curves in the main paper, we provide visual results of each model at different epochs. \Fig{ablation-vis-train} visualizes the accuracy of the reconstruction on the training set. The fitting on the training set shows high network capacity of our final model. \Fig{ablation-vis-test} visualizes the accuracy of the test set reconstruction. We early stop at epoch 1500. The result of our final model shows a good visual reconstruction compared to other models.

\begin{figure*}[t]
    \centering
    \begin{tabular}{c@{\hspace{10pt}}c@{\hspace{5pt}}c@{}c@{}c@{}c}
  
    \vspace{-0.01cm}
    \multirow{2}{*}{\includegraphics[align=c, width=0.15\textwidth, height=0.15\textwidth]{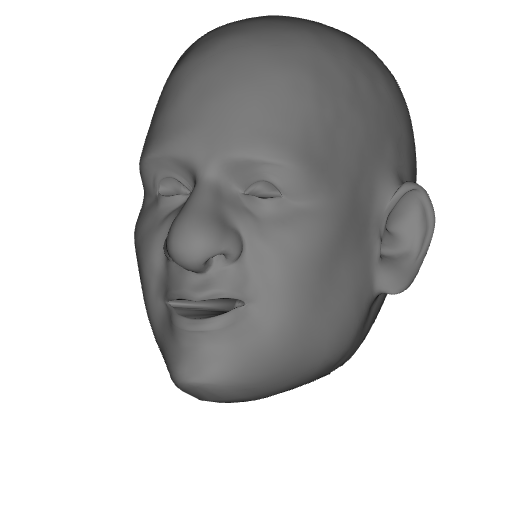}} &
    \rotatebox[origin=c]{90}{ARAP} &
    \includegraphics[align=c, width=0.15\textwidth, height=0.15\textwidth]{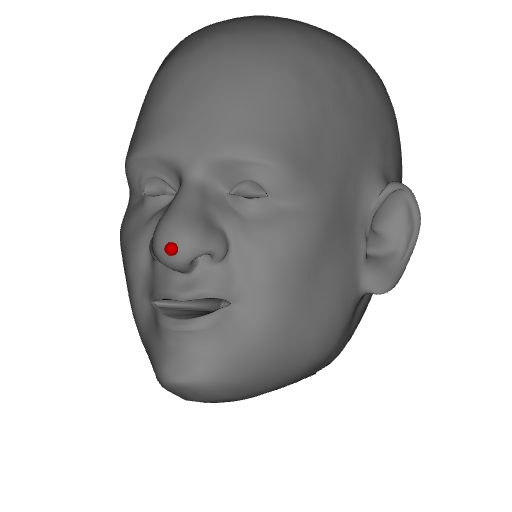} &
    \includegraphics[align=c, width=0.15\textwidth, height=0.15\textwidth]{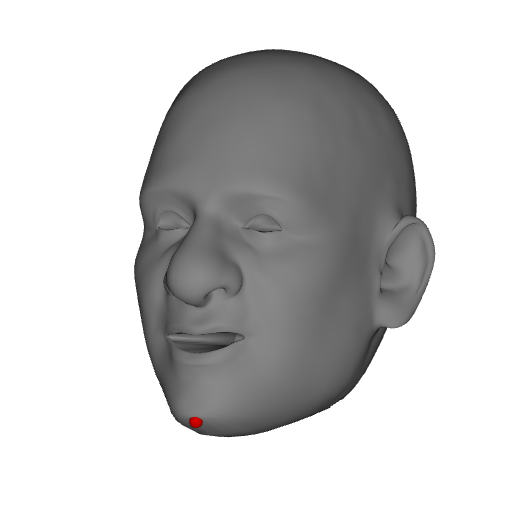} &
    \includegraphics[align=c, width=0.15\textwidth, height=0.15\textwidth]{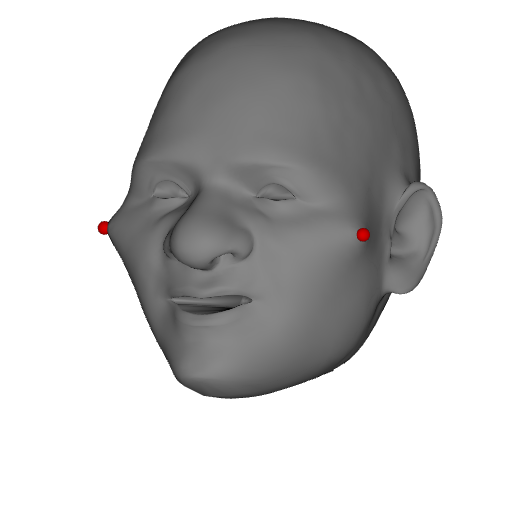} &
    \includegraphics[align=c, width=0.15\textwidth, height=0.15\textwidth]{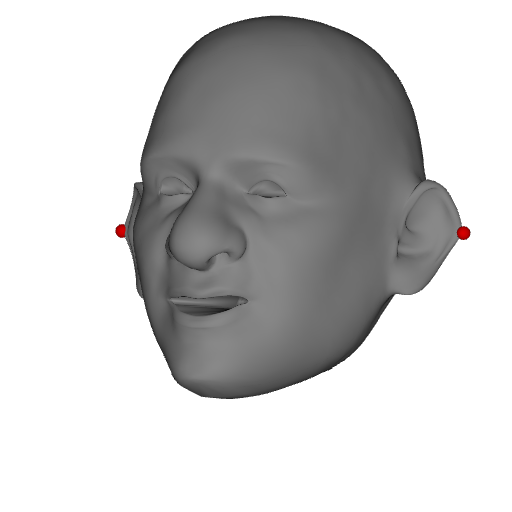}   \\
  
     &
    \rotatebox[origin=c]{90}{Ours} &
    \includegraphics[align=c, width=0.15\textwidth, height=0.15\textwidth]{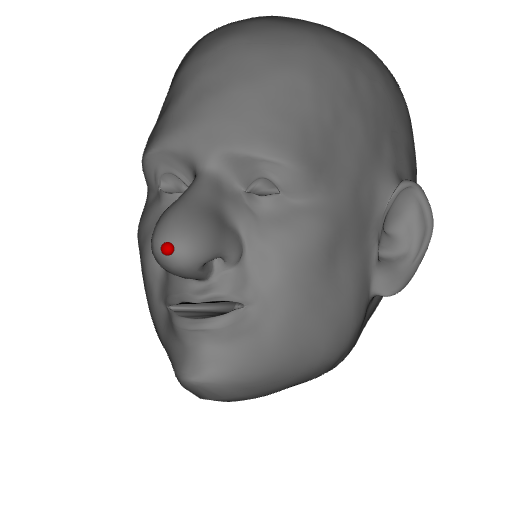} &
    \includegraphics[align=c, width=0.15\textwidth, height=0.15\textwidth]{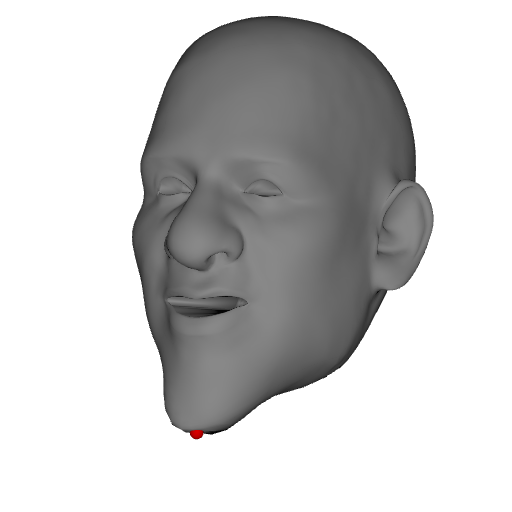} &
    \includegraphics[align=c, width=0.15\textwidth, height=0.15\textwidth]{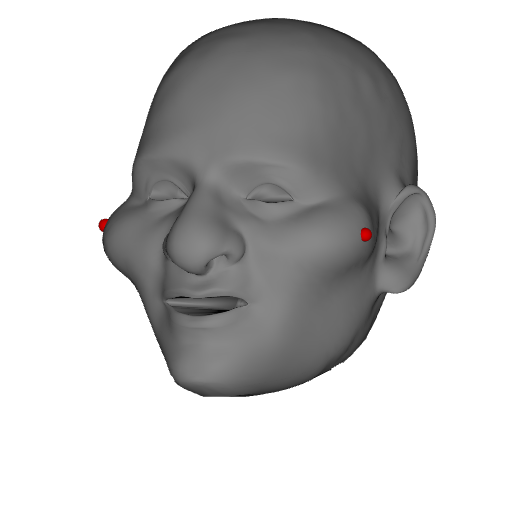} &
    \includegraphics[align=c, width=0.15\textwidth, height=0.15\textwidth]{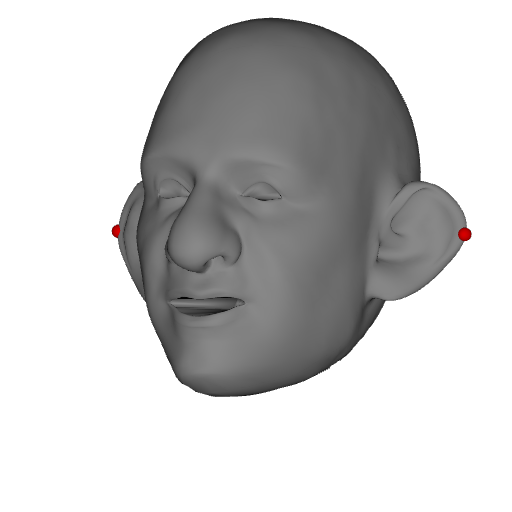}   \\
  
    Original & & (a) Nose & (b) Chin & (c) Cheek & (d) Ears \\
    \end{tabular}
    \caption{Visual comparison to ARAP deformation. Red points indicate the control points. (a,b) Since ARAP energy is insensitive to global translation by design, the optimal deformation given a single point handle displacement is naturally a global translation without any local deformation. To create local deformation using the single point in ARAP, manual selection of the region of interest (ROI) is required. Our method produces natural local deformation even with a single point constraint. (c,d) The ARAP energy increases as the area of local cells increase, so expansion or exaggeration, which we believe is crucial for caricatures, is not easily allowed in ARAP deformation. Our method naturally favors local expansion using the learned latent space.}
    \label{fig:comparison-arap}
\end{figure*}

\section{Details on 3D caricature reconstruction from 2D landmarks}
\label{sec:detail-landmark}

The 2D caricature landmarks are annotated on the silhouette. If the face pose contains large rotation, 3D landmark vertices on the 3D mesh may become hidden and not correspond to 2D landmarks. In our setting, when the input 2D caricature is not frontal, 3D landmarks on the chin may not correspond to 2D landmarks. This problem has been pointed out in \cite{wu2018alive, jiang20183d}. Their solution is to update 3D landmark vertex indices on the chin after every pose update. Similarly, we apply the silhouette landmark update algorithm after each pose estimation. 

For pose estimation, we use the 3D-to-2D point alignment algorithm presented in \cite{kemelmacher2011face}. For shape optimization, we update latent code using the same Adam optimizer with the hyperparameters we used for the training. The shape optimization is done until $(L_{t} - L_{t-1}) / L_{t}  <  1.0 \times 10^{-10}$, where $L$ is the loss value at $t$-th update.

\section{Details on InterFaceGAN in our semantic editing}
\label{sec:interfacegan}

Given semantic labels on the training set, InterFaceGAN \cite{shen2020interfacegan} enables semantic editing on the latent space given a GAN model. Although we do not use GAN architecture, we find their semantic editing is applicable to our network since our model also maps a latent vector to a face. 

We use their \textit{single attribute manipulation} technique to perform our semantic manipulation. We have semantic WebCariA \cite{ji2020unsupervised} labels for each latent vector corresponding to the 3D caricature in the training set. We find the latent direction vector for editing each semantic in the WebCariA by training a linear support vector machine (SVM) \cite{cortes1995support} for classifying each label given a latent vector. The training of SVMs is done using the WebCariA authors' published code. The normal of the hyper-plane for each SVM is the direction for editing each semantic label. We edit each face by scaling the editing vector and adding it to the source shape. The scaling factors are decided empirically.

\section{Visual comparision to ARAP deformation}
\label{sec:arap}

Handle-based deformation algorithms based on elastic energy on the surface have been extensively studied. One of the most popular of such algorithms is based on \textit{as-rigid-as-possible} (ARAP) deformation \cite{sorkine2007as}. We compare our data-driven point-handle-based deformation with ARAP deformation using the same set of constraints on the shape (\Fig{comparison-arap}). We used the ARAP implementation of libigl \cite{libigl} in the comparison. 

ARAP deformation cannot support local deformations with high stretch and area expansion that are required for face exaggeration. Without other constraints, the face generated by ARAP deformation is not guaranteed to be a valid caricature face. In contrast, our handle-based editing is constrained by the learned MLP decoder and guarantees the result to be within the latent space of caricature faces.

\section{FaceWarehouse dataset processing in automatic 3D caricature creation}

\label{sec:facewarehouse}

To train the model using both 3D caricatures and regular 3D faces, we use 3DCaricShop \cite{qiu20213dcaricshop} and FaceWarehouse \cite{cao2013facewarehouse} datasets.
As the two sets of meshes differ in mesh connectivity, we use R3DS Wrap \cite{wrap3} to register our template mesh to the mean FaceWarehouse face. Using the correspondence, we obtain the training set for regular faces in FaceWarehouse using the following procedure: 1) Randomly sample 40K points on each regular face. 2) Find the corresponding point on the FaceWarehouse mean face. 3) Find the closest point on our template mesh. Since some points on the FaceWarehouse mesh do not have a valid corresponding point to our template mesh, e.g., points on the neck, we filter out points with errors higher than 0.3. Then, we select 23,132 random points to obtain point samples for training a regular 3D face.

\bibliographystyle{ACM-Reference-Format}
\bibliography{supplementary}